	\newcommand*{\ARXIV}{}%

	\newcommand*{\COMPACT}{}%
\ifdefined\REVIEW
	\documentclass[review]{elsarticle}
\fi

\ifdefined\COMPACT
	\ifdefined\NN
		\documentclass[preprint]{elsarticle}
	\fi
	\ifdefined\ARXIV
		\documentclass[preprint, a4paper]{elsarticle}
	\fi
\fi

\ifdefined\NN
	\usepackage{lineno,hyperref}
	\modulolinenumbers[5]
\fi

\ifdefined\ARXIV
	\usepackage{hyperref}
\fi

\usepackage{graphicx}
\usepackage{bm}
\usepackage{amsmath}
\usepackage{amssymb}
\usepackage{xcolor}
\usepackage{wrapfig}
\newcommand{\phitarget}[0]{\phi^{\text{target}}}
\newcommand{\R}[1]{\in \mathbb R^{#1}}

\definecolor{darkgreen}{rgb}{0,0.5,0}
\definecolor{orange}{rgb}{1,0.5,0}
	
\newcommand*{\myrev}{\color{black}}
\newcommand*{\myreva}{\color{black}}

\usepackage{setspace}

\journal{Journal of \LaTeX\ Templates}

\ifdefined\ARXIV
	\ifdefined\COMPACT
	\fi
\fi

\begin{document}

\begin{frontmatter}

\title{Phase Portraits as Movement Primitives for Fast Humanoid Robot Control}

\author{Guilherme Maeda$^1$, Okan Koc$^2$, Jun Morimoto$^{1*}$}
\address{g.maeda@atr.jp, okan.koc@tuebingen.mpg.de, xmorimo@atrj.jp}
\address{$^{1}$ATR Computational Neuroscience Laboratories. Kyoto, Japan}
\address{$^{2}$Max Planck Institute. T{\"u}bingen, Germany}

\ifdefined\NN
	\cortext[mycorrespondingauthor]{Corresponding author: xmorimo@atr.jp.
		ATR. Department of Brain Robot Interface.
		2-2-2 Hikaridai, Seika-cho, Soraku-gun, Kyoto 619-0288, Japan} %\ead{}

\fi

\setstretch{1.2}    
\begin{abstract}
	Currently, usual approaches for fast robot control are largely reliant on solving online optimal control problems. 
	Such methods are known to be computationally intensive and sensitive to model accuracy. On the other hand, animals plan complex motor actions not only fast but seemingly with little effort even on unseen tasks. This natural sense to infer temporal dynamics and coordination motivates us to approach robot control from a motor skill learning perspective to design fast and computationally light controllers that can be learned autonomously by the robot under mild modeling assumptions. This article introduces Phase Portrait Movement Primitives (PPMP), a primitive that predicts dynamics on a low dimensional phase space which in turn is used to govern the high dimensional kinematics of the task. The stark difference with other primitive formulations is a built-in mechanism for phase prediction in the form of coupled oscillators that replaces model-based state estimators such as Kalman filters. The policy is trained by optimizing the parameters of the oscillators whose output is connected to a kinematic distribution in the form of a phase portrait. The drastic reduction in dimensionality allows us to efficiently train and execute PPMPs on a real human-sized, dual-arm humanoid upper body on a task involving 20 degrees-of-freedom. We demonstrate PPMPs in interactions requiring fast reactions times while generating anticipative pose adaptation in both discrete and cyclic tasks.
\end{abstract}

\begin{keyword}
	Imitation learning, reinforcement learning, movement primitives, phase estimation, coupled oscillators.
\end{keyword}

\end{frontmatter}

\ifdefined\NN
	\linenumbers
\fi

\ifdefined\REVIEW
	\setstretch{1.5}
	\fi
	\ifdefined\COMPACT
	\setstretch{1.00}
\fi

\section{Introduction}

\begin{figure}
\centering
\includegraphics[width=0.95\textwidth]{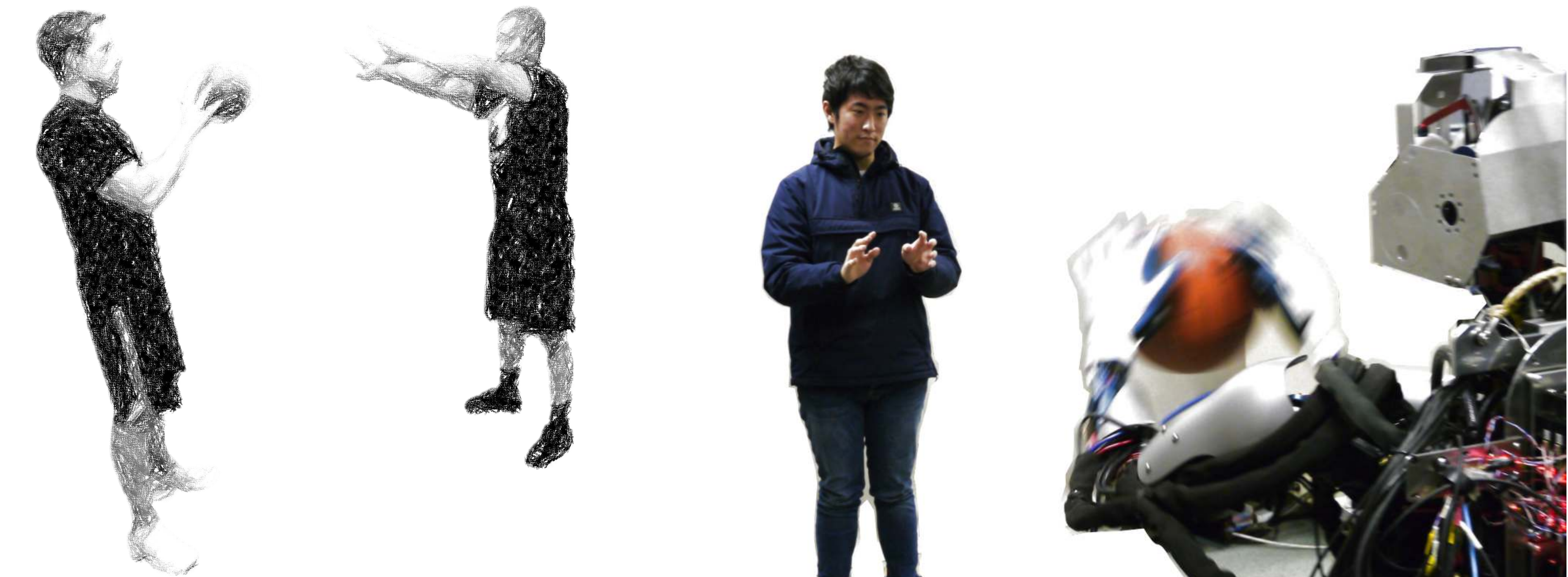}
\caption{ \textbf{Playing with a ball.} Humans are naturally endowed with the ability to estimate the time-to-contact while coordinating multiple degrees of freedom even without knowing the exact dynamics of the flight of the ball and the involved contact forces.    
	To endow robots with such capabilities, this article introduces Phase Portrait Movement Primitives (PPMPs).
	A PPMP is capable of estimating the future temporal states of an interaction by using coupled oscillators as a low-dimensional dynamical model in phase space.
	By using policy search, the robot learns to combine the estimated phase with the high-dimensional kinematics of the task represented as a phase portrait.
}
\label{fig:kids}
\end{figure}

\noindent
Humans react to changes in the environment by adapting and coordinating complex motor actions gracefully and skillfully.
Consider motor skills that basketball players exhibit when passing and receiving a ball as shown in Figure \ref{fig:kids}.
They naturally coordinate the position of the hands according to the progress of the ball trajectory, quickly adapting both in time and space.
Yet, in contrast to the heavy online optimization typically found in robot control  (e.g. \cite{baumlKinematicallyOptimalCatching2010,erezIntegratedSystemRealtime2013,kim2014catching,ardakaniRealtimeTrajectoryGeneration2015}), 
no conscious effort seems to be spent on either the ball prediction or the trajectory planning of the hands.
This natural agility could be in part because, from an early age, humans seem to develop a sense of timing\cite{lee1980_visuomotor_timetocontact} that proves essential in fast motor actions.
Moreover, the fact that we continuously anticipate the interaction by adapting our pose---e.g. when someone tricks us by rocking a ball towards us, pretending a pass but not doing it---indicates that this sense of timing is given by some predictive mechanism that makes us anticipate poses {\myreva(i.e. our kinematic configuration)} according to the expected progress {\myreva(i.e. the phase)} of the interaction.

Animal agility also seems to be related to the existence of motor primitives as internal models learned and improved from experience.
These primitives provide a mapping from sensory signals to motor commands \cite{wolpert1998multiple,thoroughman2000learning,wolpert2011principles} coordinating the relevant degrees-of-freedom in the system and can be quickly queried to achieve fast adaptation.
In robot control, researchers have for many years tried to create the corresponding counterparts as movement primitives acquired via imitation \cite{schaalImitationLearningRoute1999}.
While the search space decreases considerably by using the templates provided by primitives, its reliance on future state prediction is still a challenge for fast reactions.
For example, Kober et al. \cite{kober2010movement} designed a  specialized Kalman filter to predict the future states of an incoming ball in robotic table tennis.
That work used Dynamical Movement Primitives (DMPs) \cite{ijspeert2013dynamical}, a representation widely used within the robot learning community.
A goal of our method is to incorporate a predictive sense of timing (such as the time-of-contact of a flying object) as an intrinsic part of the primitive without relying on the detailed physics models required by traditional state estimators such as Kalman filters and state observers.

\begin{figure}    
\centering
\includegraphics[width=0.99\linewidth]{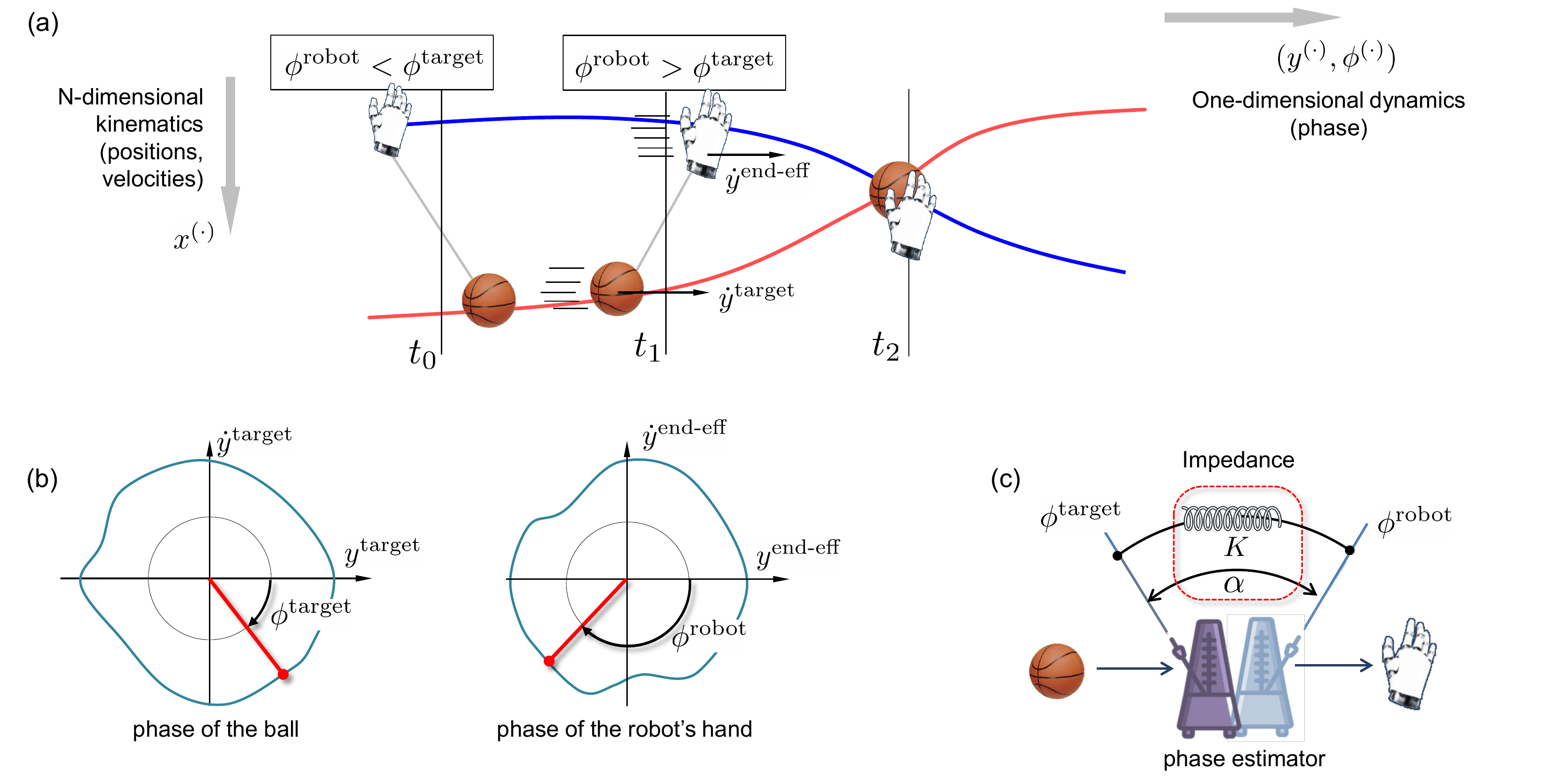}
\vspace{-.005cm}
\caption{
	\textbf{Phase prediction and adaptation for ball catching.} 
	{\myrev
		(a) An illustration of a motivating problem where a robot must continuously observe the flight of a ball and adapt its phase $\phi^\text{robot}$ to the ball phase ${\phi}^{\text{target}}$ such that a successful catching is possible.
		Pure reactive feedback tracking on the ball position can not account for the fact that the robot must advance its phase, as shown at time $t_1$, to preemptively prepare for the ball landing at time $t_2$.
		(b) The phase is defined as the phase-plane angle. 
		In this particular example, the phase is computed using only the $y$ coordinates of the movements since this is the direction related to the progress of the task (as opposed to $x$).
		(c) We hypothesize that the dynamics of such predictive tasks can be approximated by two coupled oscillators interacting via an impedance.
		The impedance adapts the robot's progress according to the target's progress.        
		At runtime,
		the robot acts preemptively when the relation between phases leads to ${\phi}^{\text{robot}}>\phi^{\text{target}}$ such as in $t_1$.
	}
}
\label{fig:introa}
\end{figure}

A difficult requirement in tasks involving fast dynamics is that robot control commands must account for the future states of the interaction while being computed on a very limited time budget, often at the order of tens of Hertz.
In {\myreva Figure \ref{fig:introa}(a)}, this difficulty is cartooned by a robot that when observing the flight of a ball at time $t_0$, must advance its progress as shown at time $t_1$ such that it can compensate for the many sources of delays (inertia, torque limits, friction, backlash, etc.) when preparing for the landing of the ball at $t_{\text{2}}$.
Here, we refer to the robot's and ball's progresses by their phases, ${\phi}^{\text{robot}}$ and $\phi^{\text{target}}$, respectively.
{\myrev We define phase as a particular distinct stage within a series of stages defining a task.
Thus, when ${\phi}^{\text{robot}} > \phi^{\text{target}}$ the robot is acting in an anticipative manner.
Quantitatively, in this article, the phase used is obtained as the angle of the current state of the dynamical system on the phase plane as shown in Figure \ref{fig:introa}(b).
(Refer to \cite{slotineAppliedNonlinearControl1991} Chap. 2 for background on phase plane analysis).
}

Using this illustrated case as motivation, in the broadest possible terms, two principal directions to solve this problem are to use models that can be exploited by online optimization or to rely on trial-and-error under the framework of model-free reinforcement learning.

Within the class of methods based on models and online optimization, for many years, Model Predictive Control (MPC) has been considered the gold standard in fields such as chemical processes and more recently in robotics \cite{garciaModelPredictiveControl1989, erezIntegratedSystemRealtime2013, ardakaniRealtimeTrajectoryGeneration2015}.
MPC is characterized by the computation of a look-ahead trajectory (whose length is defined by the horizon) which starts at the current state of the task.
This trajectory is executed partially until a new look-ahead trajectory is ready to replace the previous one.
By constantly refreshing the reference trajectory, the effects of disturbances and modeling errors are greatly reduced as they are continuously absorbed within the states used to generate the subsequent trajectory.
As a result, MPC---and online optimal control, in general---not only relies on the designer's domain knowledge 
but also on fast optimization routines and powerful computing (e.g. \cite{baumlKinematicallyOptimalCatching2010, ardakaniRealtimeTrajectoryGeneration2015, marcucciApproximateHybridModel2017}).
This reliance on models and computational requirements makes such approaches less suited to fulfill the vision of agile robots capable of learning tasks autonomously.
On the other spectrum, reinforcement learning is an extremely general approach for robotic autonomous learning\cite{koberReinforcementLearningRobotics2013Z}.
However, the lack of model information leads to an unrealistic amount of experience to be acquired via trial-and-error; 
reducing its practicality to low dimensional problems.

{\myrev
We propose a middle-ground solution in terms of model requirements which we will loosely refer to as \emph{semi-model-free}.
Research on robot motor skill learning provides evidence that the time and coordination found in animals can be represented by movement primitives and efficiently optimized due to its low dimensionality (e.g. \cite{ijspeert2007swimming,morimotoBiologicallyInspiredBiped2008, mulling2011biomimetic}).
Following the skill learning approach, we implement and investigate whether the \emph{dynamics} of the task can be sufficiently 
approximated by a general oscillator model (e.g. \cite{strogatz2001nonlinear}) that has also been adopted in the locomotion literature.
Under a manipulation task such as the ball catching, we hypothesize that the dynamics of the interaction between the robot and an external agent can be approximated by the dynamics of two oscillators coupled by an impedance (details in the next section).
By construction, we enforce that the \emph{kinematics} of the task is modeled independently from the dynamics of the phase.
This kinematic model leverages human demonstrations to learn a probabilistic phase portrait that covers the vicinity of the observed trajectories.
The phase dynamics and the kinematic representations do not rely on domain knowledge or parameter system identification required by model-based methods.
The dimensionality of the whole system decreases to a degree that is amenable to the use of policy search reinforcement learning methods.
}

The main \emph{contribution} of this article is to propose a control method based on movement primitives with an intrinsic mechanism for fast phase estimation.
As the primitive is represented as a phase portrait we name the method Phase Portrait Movement Primitives (PPMP).
This close connection between phase estimation and primitives allows for first solving the \emph{dynamics} of the interaction in the phase space, subsequently reducing \emph{pose coordination} to a kinematic problem.
The benefit is that at runtime, PPMPs have negligible computational {\myreva load, allowing} for fast generation of predictive actions.
PPMP is a general method devised to support the vision of future robots capable of learning tasks completely autonomously.
This article empirically evaluates the sufficiency of this representation and its efficiency for reinforcement learning by applying it on a variety of tasks; from dynamic manipulation of a flying ball, to handovers with a human, to footstep placing for walking.
Our preliminary investigations on the phase oscillator dynamics which led to the PPMP proposed in this article first appeared as a short conference paper \cite{maedaReinforcementLearningPhase2018}. 
This article has a mature view of the method, in which we more deeply analyze its fundamental components while guiding the reader through the synthesis of PPMPs.
In addition to formalizing the concept of phase portraits for the first time, this article also extends experimental cases significantly, comparing and discussing the parallels with other primitive formulations.

\section{Coupled Oscillators and Phase Portraits}

\noindent
This section introduces the main components of PPMPs: a phase estimator in the form of coupled phase oscillators used for predicting the dynamics of the interaction, and a probabilistic phase portrait as a distribution of demonstrated trajectories for kinematic coordination.
{\myreva Because the connection between phase oscillators (dynamics) and phase portraits (kinematics) are not observable during humans demonstrations, and in fact modeled independently,
the last step of PPMP consists of merging the oscillators and the phase portrait as a single policy via reinforcement learning (RL).}

\subsection{Timing the Dynamics of Interactions via Coupled Oscillator} \label{sec:coupled_osc}

{\myrev
Oscillators have been used as computational models of animal skill coordination to provide robots with a central-pattern generator for locomotion \cite{ijspeert2007swimming}, posture control \cite{melnyk2018synergistic}, and upper limb \cite{jouaitiCPGbasedControllersCan2018} control.
We revisit a model based on coupled oscillators \cite{cohen1982nature, strogatz2001nonlinear, morimotoBiologicallyInspiredBiped2008, jouaitiCPGbasedControllersCan2018} and reframe it as a predictive model to estimate temporal dynamics in manipulation tasks.
If properly tuned through learning iterations, our expectation is that the oscillators allow the robot to {\myrev dynamically adapt its} actions as a function of the evolution of an external agent.

Figure \ref{fig:introa}(c) shows the concept of the phase mechanism adopted in this work as two metronomes coupled by an impedance.
This controller emulates an interacting stiffness between the metronomes with a phase difference.
The interaction between the oscillators can be expressed as
\begin{equation}
\dot{\phi}^{\text{robot}} = \omega + K \sin (  \phi^{\text{target}} - \phi^{\text{robot}} + \alpha ),
\label{eq:coupled_osc}
\end{equation}
where $\phi^{\text{target}}$ is the phase of the target (e.g. the flying ball),
$\phi^{\text{robot}}$ is the phase of the robot,
$K$ is a tunable stiffness that defines how strictly the phase of the robot tracks the phase of the target,
$\alpha$ is a tunable phase difference between the oscillators,
and 
$\omega$ is the tunable natural frequency of the system, that is, the velocity at which the robot would move if no correction was applied.

In control terms, $\phi^{\text{target}}$ is the reference input, $\phi^{\text{robot}}$ is the output variable, and $\alpha$ is an offset.
%If the offset is zero and the phase difference converges to zero, then the robot and the target will be moving side-by-side on the temporal dimension, such as time $t_2$ in Figure \ref{fig:introa}(a).
Appropriate parameter values in \eqref{eq:coupled_osc} can be used to modulate the robot characteristics with respect to the target.
For example, to make the robot behave in an anticipative manner in relation to the movement of a flying ball---a situation illustrated in Figure \ref{fig:introa}(a) at time $t_1$ when ${\phi}^{\text{robot}} > \phi^{\text{target}}$.
%While the dynamics \eqref{eq:coupled_osc} is a feedback control law which is \emph{reactive} in nature, the effect of its output in the task is \emph{predictive} when $\alpha > 0$ or when $\omega > 0$.
Phase prediction using coupled oscillators is computationally very light as is it only involves integrating \eqref{eq:coupled_osc}.
}

\subsection{Kinematic Coordination via Probabilistic Phase Portrait} \label{sec:PPP}

The phase is related to the {\myreva dynamical} dimension of the task---represented by the horizontal axis in Figure \ref{fig:introa}---and does not contain the {\myreva kinematic information} such as the joint configurations of the robot.
Here, we describe how the joint positions of the agent can be learned and computed as a function of the target's position via probabilistic inference.
A probabilistic approach for fast robot pose adaptation has two main benefits.
First, by assuming a normal distribution, a closed-form solution can be used to instantaneously infer robot poses given observed context positions.
Second, the phase portrait can be designed from demonstrations, rendering a methodology that requires little domain knowledge.

As an imitation learning approach, the idea is to obtain a $N$ number of variations of paired trajectories of duration $T$ containing robot joint angles $\bm q^{\text{demo}}_{1:T}$ and target movements in Cartesian space (e.g. via motion capture) $\bm x^{\text{demo}}_{1:T}$ with
$\bm{x}^{(\cdot)}_t = (x^{(\cdot)}_t, y^{(\cdot)}_t, z^{(\cdot)}_t)$
from multiple demonstrations.
Each $n$-th paired trajectory is assumed to be a sample from a joint distribution 
$(\bm q^{\text{}}, \bm x^{\text{}})_{1:T} \sim P(\bm q, \bm x)_{1:T}$.
Assuming normal distributions, at each time step $t \in \{1,..,T\}$, we can write

\begin{equation}
P(\bm{q}, \bm{x}  )_{t} =  \mathcal N (\{ \bm\mu_{q, x} \}_{t}, \{ \bm\Sigma_{q, x} \}_{t} ),
\label{eq:Pxq}
\end{equation}
where
\begin{equation}
\{ \bm\mu_{q, x} \}_t= 
\begin{bmatrix} 
\{ \bm\mu_{q} \}_t, \ \ \{ \bm\mu_{x} \}_t
\end{bmatrix}^\top = 
\text{mean}(\{(\bm q,\bm x)_{1:N}\}_{t} ) \quad  \text{ and}
\label{eq:mu}
\end{equation}
\begin{equation}
\{  \bm\Sigma_{q,x}  \}_t= 
\begin{bmatrix}
\{ \bm\Sigma_{qq} \}_{t}  &  \{  \bm\Sigma_{qx}     \}_{t} \\
\{ \bm\Sigma_{xq} \}_{t}  &  \{  \bm\Sigma_{xx}    \}_{t}
\end{bmatrix} = 
\text{cov}(\{(\bm q,\bm x)_{1:N}\}_{t} ),
\end{equation}
where the interval ${1:N}$ shown in the subscript indicates all $N$ demonstrations at the time step  $t$.

During execution, at each instant $t$, a target observation $\bm{x}_t = \{x_t,y_t,z_t\}$ is made and the corresponding joint configuration of the robot is computed as a conditional distribution
\begin{equation}
P(\bm{q} | \bm{x}^{\text{}})_{t} = \mathcal N (  \{ \bm\mu_{q|x}  \}_{t},  \{ \bm\Sigma_{q|x} \}_{t} ),
\end{equation}    
with
\begin{equation}
\begin{split}
\bm\mu_{q|x} & = \bm\mu_{q} + \bm\Sigma_{qx} \bm\Sigma_{xx}^{-1} (   \bm{x}^{} - \bm\mu_{x}  ) \\
\bm\Sigma_{q|x} & = \bm\Sigma_{qq} - \bm\Sigma_{qx} \bm\Sigma_{xx}^{-1} \bm\Sigma_{xq},
\end{split}
\label{eq:inference}
\end{equation}
where the subscript $t$ was dropped.

\begin{figure*}    
\centering
\includegraphics[width=0.9995\linewidth]{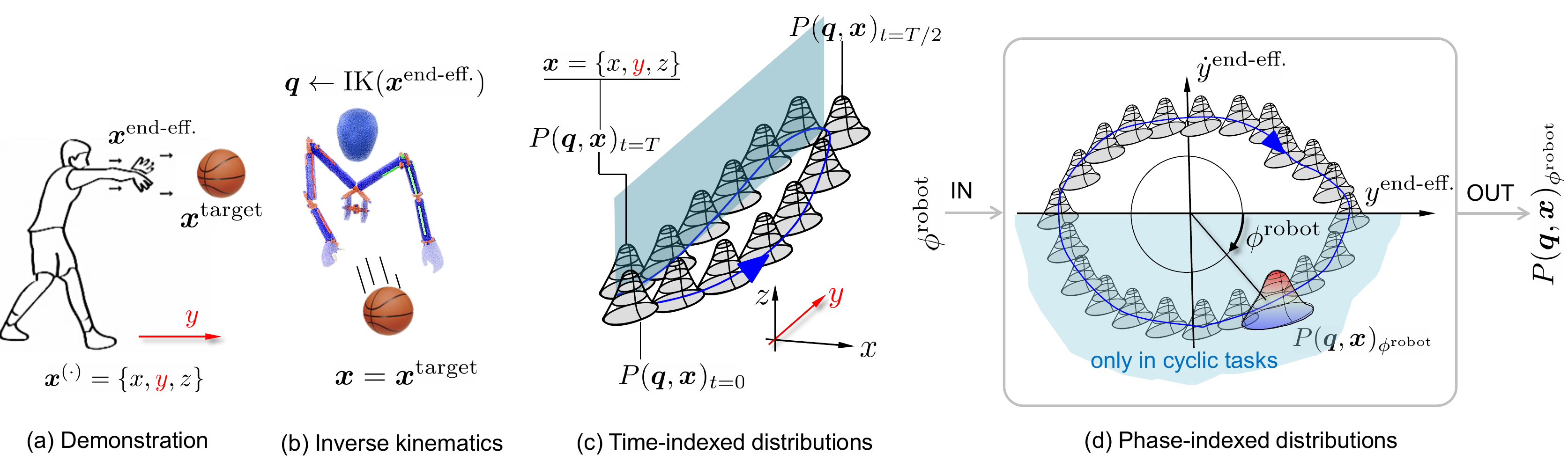}
\vspace{-.5cm}
\caption{
	\textbf{Representing Kinematics as a Probabilistic Phase Portrait.} 
	The Phase Portrait Movement Primitives (PPMP) consists of a sequence of joint distributions that correlates demonstrator and target positions.
	These joint distributions are first captured via demonstration (a) and transformed into joint angles via inverse kinematics (b).
	The trajectory of distributions (c) are then laid on the phase-plane of the end-effector ($y^{\text{end-eff.}}, {\dot y}^{\text{end-eff.}}$). 
	The distributions can then be accessed by the phase angle $\phi^{\text{robot}}$.
	The semi-transparent plane in (c-d) shows the only distinction between cyclic and single-stroke tasks, wherein the latter case only half of the distributions are present as no path of return exists.
	The y-axis is highlighted as the axis used to construct the phase plane as it is the direction associated with the progress of the task. 
}
\label{fig:figure1c}
\end{figure*}

In practice, many tasks cannot be demonstrated directly using a robot either because most robots are not backdriveable (no kinesthetic teaching capability) or because the task is too fast to be executed while moving a robotic arm (like catching a flying ball).
As shown in Figure \ref{fig:figure1c}(a), a more natural approach is to observe a human executing the task while interacting with a target where paired hands and target trajectories are recorded.
Later, the robot trajectory is computed via inverse kinematics $\bm q^{\text{demo}}_{1:T} \leftarrow \bm x^{\text{end-eff.}}_{1:T}$ as shown in Figure \ref{fig:figure1c}(b).

Figure \ref{fig:figure1c}(c) shows the sequence of joint distributions \eqref{eq:Pxq} in time-domain.
The plane cutting the trajectories in half is used to indicate the difference between cyclic and single-stroke tasks.
In the former, the distributions return close to the initial state, and in the latter, only the first half of the distributions exist.
PPMPs apply seamlessly in both cases.

To connect the oscillator and the phase portrait consistently, we re-index the temporal sequence of kinematic distributions with the phase of the end-effector motion such that the angle of the phase plane becomes the input 
% gjm
obtained 
from the oscillator.
This indexing, illustrated in Figure \ref{fig:figure1c}(d), uses the mean value of the end-effector trajectory to compute the phase of the robot as the angle 
\begin{equation}
\phi^{\text{robot}} = \phi_{t}^{} = -\arctan  ( \dot{y}^{\text{end-eff.}}_{t},  y^{\text{end-eff.}}_{t}   ), \ t \in [1, \ T], y^{\text{end-eff.}}_{t} \in \mathbb{R}.
\label{eq:phaserobot}
\end{equation}
The primitive in Figure \ref{fig:figure1c}(d) is a sequence of distributions on the phase plane and can be interpreted as a probabilistic phase portrait.
For online execution, the time-based phase $\phi_{t}$ is replaced by the phase computed by the oscillator $\phi^{\text{robot}}$.
%
% gjm: Jun's comment to address
%**************
%
% gjm: modified the sentence a bit
% The idea is to use the phase $\phi^{\text{robot}}$ to retrieve $P(\bm{q}, \bm{x}  )_{\phi}$ from the phase portrait. 
This re-indexing of distributions allows us to retrieve $P(\bm{q}, \bm{x}  )_{\phi}$ from the phase portrait using the phase $\phi^{\text{robot}}$ instead of time. 
%
%
% gjm: move this sentence from here (search for gjm1 within the LaTeX text)
%This distribution is then conditioned on the observed target to produce a set of joint angles\footnote{Alternatively, one could retrieve $\bm x^{\text{end-eff.}}_{\phi}$ from the phase portrait and use inverse kinematics to compute a solution for the inferred end-effector position online. In principle, the only compelling reason for doing so is when the true distribution cannot be captured as $\mathcal N (\{ \bm\mu_{q, x} \}_{t}, \{ \bm\Sigma_{q, x} \}_{t} )$.} $\bm{q} \sim P(\bm{q}|\bm{x}  )_{\phi}$.
Note that in the single stroke case, the phase portrait only covers only the first two quadrants of the phase plane.

{\myrev 
Note that the coordinate chosen to compute the phase in \eqref{eq:phaserobot} (in this case the Y direction) is a designer's choice.
Currently, the method assumes that the reference frame and the coordinate to which to compute the phase of the task are given.
For a variety of tasks, it seems intuitive to define this coordinate by a vector normal to the frontal plane of the body, parallel to the sagittal plane of the robot. 
The heuristic holds because this is the direction that provides the minimum distance when the robot is facing an incoming target.
This distance, accounts for the progress of the task, as opposed to lateral displacements between the target and the robot.
We used the same heuristic on all experiments in the article, namely, ball pushing, handovers, and leg control.
In this article, we did not investigate methods to extract the optimal projection automatically, which would potentially involve the 3-D analysis of the movement and its relevant components, for example, via principal component analysis.
}

\subsection{Consolidating Oscillators and Phase Portraits as a Single Policy} \label{sec:rl}

\begin{figure*}    
\centering
\includegraphics[scale=0.5]{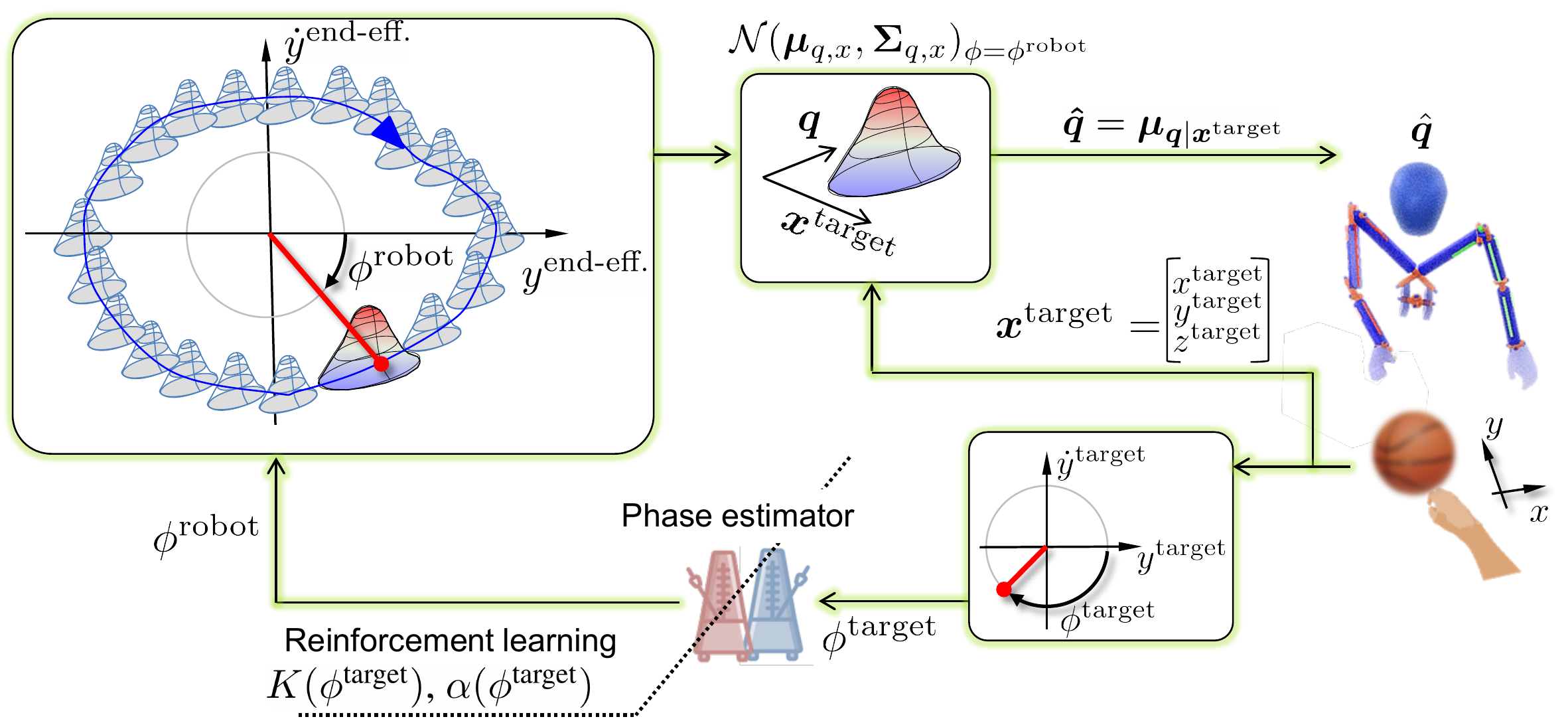}
\vspace{-.005cm}
\caption{
	\textbf{Learning to combine coupled oscillators with the probabilistic phase portrait as a control policy.} 
	The full method where the oscillator output selects a joint distribution on the phase portrait. This distribution is then conditioned on the target position to retrieve robot joint angles. The use of policy search RL on the parameters coupling the oscillators allows the robot to consolidate the nonlinear synchronization between the oscillator dynamics and robot poses as a single and consistent policy. 
}
\label{fig:RL}
\end{figure*}

So far, the phase estimation resulting from the dynamics of the coupled oscillators is unrelated to the kinematics of the phase portrait.
Not only the oscillators and the portraits represent different policies (dynamic and kinematic) but they have been designed independently in the previous sections.
Indeed, the relation between oscillators and phase portraits cannot be observed during the demonstration.
Another issue is that we want the robot to adapt to a variety of situations for which demonstrations---and their implicit dynamics---are not available. 
Thus, it is necessary to consolidate these two components as a single robust dynamical policy.
To this end, we propose the scheme in Figure \ref{fig:RL} where the dynamics between the coupled oscillators governs the selection of distributions on the phase portraits.
%
% gjm1: Jun comment add this line
The selected normal distribution is conditioned on the current target observation $\bm x^\text{target}$.
The corresponding estimated robot's pose is obtained from the resulting mean ${ \bm {\hat{q}}} =\bm \mu_{\bm q| \bm x^{\text{target}}}$ which can be fed as reference angle values to the robot's joint controller.

Referring to \eqref{eq:coupled_osc}, the only open parameters that allow this consolidation are the coupling elements, stiffness $K$ and phase shift $\alpha$, between oscillators.
The natural frequency $\omega$ governs the velocity of the cycle when the robot is free from external inputs.
For manipulation tasks, we assume that the robot should not move when the target does not move, thus we use $\omega=0$ as the default value for all the experiments in this article.
Fixed values of $K$ and $\alpha$ generate a linear control parametrization, which, except for simple single stroke tasks, show insufficient to represent different task regimes.
Thus, to allow for a nonlinear parametrization of oscillator dynamics we define the coupling parameters as two functions $K = K(\phitarget)$ and  $\alpha = \alpha(\phitarget)$.
Finding the optimal PPMP policy means finding the shapes of $K(\phitarget), \  \alpha(\phitarget)$ such that a cost is minimal when the task is finished successfully.

Except for simple systems, this optimization is usually not suited for model-based methods as the interaction dynamics are the result of many unknown lumped effects such as the delays in the robot actuation, mechanism nonlinearities (e.g. friction and backlash), bi-manual asymmetric exchange of contact forces between the hands and the manipulated object, inaccuracies of the perception system and its delays, etc.
Thus, for the general case, 
a gradient-free direct policy search approach \cite{deisenroth2013survey} is better suited as it optimizes the functions $K(\phitarget), \  \alpha(\phitarget)$ directly from rewards.

As usual in policy search, the optimization is done on weights as parameters that control the amplitude $N$ radial-basis functions (RBFs)\footnote{An alternative to RBFs is to use circular features such as von Mises bases to ensure the continuity of $K(\phitarget), \  \alpha(\phitarget)$ when  $\phitarget$ wraps around a full cycle. We did not notice the continuity to be important as $\phi^{\text{robot}}$ results from the integration of a differential equation. Thus, even if the parameters are discontinuous, the phase trajectory that results from using these parameters is smooth by construction.}. 
For a fixed set of basis functions 
\begin{equation}
	[K(\phitarget), \, \alpha(\phitarget)] = [ \psi_1(\phitarget) \ ... \   \psi_N(\phitarget)  ] [\bm w_K, \, \bm w_\alpha], 
	\label{eq:func}
\end{equation}
where $\bm w_K  \R{N}$ and  $\bm w_\alpha \R{N}$ are the weight vectors.
The phase of the target $\phitarget$ is defined over the interval $[-\pi, \ \pi]$ for cyclic tasks and  $[0, \ \pi]$ for single-stroke tasks.
The basis functions $\psi_{1:N}(\phitarget)$ have the position of their centers uniformly distributed within the $\phitarget$ range.

%The definition of the number of basis functions is usually an empirical process.
%Lower $N$ values impose smoother functions while larger values allow for functions requiring rapid transitions.
%In our experiments, a typical value was $N=10$.  
%% gjm
%%The weights can be computed by solving the normal equations (refer to \cite{bishop2006pattern} chap. 3 for detailed procedure)
%The weights $\bm w_K$, $\bm w_{\alpha}$ are the respective projections of $K$ and $\alpha$ on the subspace spanned by the vectors in $\mathbf \Phi_{}$ and can be computed by solving the normal equations (refer to \cite{bishop2006pattern} chap. 3 for detailed procedure)
%\begin{equation}
%\begin{split}
%\bm w_K       &  = 
%\left(    \mathbf \Phi^\top_{}   \mathbf \Phi_{}  \right)^{-1}  \mathbf \Phi^\top_{} K_{} (\phi^{\text{target}}_{-\pi:\pi})\\
%\bm w_{\alpha} & = 
%\left(    \mathbf \Phi^\top_{}   \mathbf \Phi_{}  \right)^{-1} \mathbf \Phi^\top_{} \alpha_{} (\phi^{\text{target}}_{-\pi:\pi}).
%\end{split}
%\end{equation}

% gjm linking 9 and 10
The goal of the policy search is to simultaneously optimize the functions $K(\phitarget)$ and $\alpha(\phitarget)$ by optimizing the concatenated vector $\bm w = [\bm w_K^\top, \, \bm w_\alpha^\top]^\top \R{2N}$ such that a cost criterion $C(\cdot)$ designed for the task at hand is decreased.
Under the episodic case, the general workflow consists of creating a total of $R$ number of  roll-out variations, evaluating the cost of each roll-out, and updating the weight vector based on these variations. 
This process is then repeated until convergence or a specified criterion (e.g. maximum number of weight updates) is achieved.

The $r$th roll-out variation is obtained by perturbing the weight vector with additive white Gaussian noise $\bm w_{r}  = \bm w +  \bm \epsilon_r$ where $\bm \epsilon_r \sim \mathcal N(0, \bm{\Sigma}_n)$  and $\bm{\Sigma}_n$ is the noise variance.
The execution of the perturbed policy under the weights $\bm w_{r}$ results in a trajectory $\bm{\tau}_r$ with a cost $C(\bm{\tau}_r)$. 
While the policy search literature provides numerous ways to update the weights, here we follow the  Pi$^\text{BB}$~\cite{stulp2012policy} methodology, which is a black-box optimization\footnote{PPMPs do not require a particular policy search algorithm. In principle, it can be used with any black-box optimizer.} 
algorithm based on Pi$^2$ \cite{theodorou2010reinforcement}.
The updates are of the form
\begin{equation}
	\bm w^{\text{new}} \leftarrow \bm w^{\text{old}} + \sum_{r=1}^{R}[P(\bm {\tau}_r) \bm{ \epsilon }_r],
\end{equation}
where the old parameters $\bm w^{\text{old}}$ are added with the weighted average of the exploration noise $\bm \epsilon_r$ (i.e. the expected noise).
The probability weight of each roll-out is computed with    
\begin{equation}
	P(\bm{\tau}_r) = \frac{  \exp[-\lambda C(\bm{\tau}_{r})]    }  {\sum_{r'=1}^R \exp[-\lambda C(\bm{\tau}_{r'})] },
\end{equation}
where $\lambda$ is a constant scalar that modulates the amount of discrimination among the costs in the batch.

%
%As the policy search algorithm we adopted Pi$^\text{BB}$~\cite{stulp2012policy}    which is a black-box optimization\footnote{PPMPs do not require a particular policy search algorithm. In principle, it can be used with any black-box optimizer.} 
%algorithm based on Pi$^2$ \cite{theodorou2010reinforcement} to search for the optimal vector $\bm w = [\bm w_K^\top, \bm w_\alpha^\top]$. 
%The updates are of the form
%\begin{equation}
%	\bm w^{\text{new}} \leftarrow \bm w^{\text{old}} + \sum_{r=1}^{R}[P(\bm {\tau}_r) \bm{ \epsilon }_r],
%\end{equation}
%where a parameter update is done every $R$ batches of roll-outs.
%The old parameters $\bm w^{\text{old}}$ are updated by using the weighted average of the exploration noise $\bm \epsilon_r  \R{N}$, which are sampled from a zero-mean Gaussian distribution.
%The weight of each roll-out is computed with    
%\begin{equation}
%	P(\bm{\tau}_r) = \frac{  \exp[-\lambda C(\bm{x}_{r})]    }  {\sum_{r'=1}^R \exp[-\lambda C(\bm{x}_{r'})] },
%\end{equation}

While we motivate the policy search as a way to find an optimal policy, an alternative interpretation is to view the optimization as a parameter estimation problem via RL \cite{Morimoto2007RLS} since the coupling between oscillators is a parameterized control law.

%For the particular scheme proposed in Figure \ref{fig:RL}, the execution of a roll-out consists in using the exploratory weights to retrieve corresponding functions $K(\phitarget)$ and  $\alpha(\phitarget)$ with         \eqref{eq:func}.
For the particular scheme proposed in Figure \ref{fig:RL}, the initialization of a roll-out consists in retrieving the corresponding functions $K(\phitarget)$ and  $\alpha(\phitarget)$ from a noisy vector $\bm w_r$ with \eqref{eq:func}.
During execution and under a continuous stream of target observations $\bm x^\text{target}$, the target phase $\phi^\text{target}$ is computed and used as the input for the coupled oscillators, from which the robot phase is found as
\begin{equation}
\dot{\phi}^{\text{robot}} = \omega + K(\phitarget) \sin (  \phi^{\text{target}} - \phi^{\text{robot}} + \alpha(\phitarget) ).
\label{eq:coupled_osc_learning}
\end{equation}
The phase of the robot indicates which normal joint distribution from the phase portrait to be used.
This distribution is conditioned on $\bm x^\text{target}$ to obtain the estimated joint configurations of the robot as ${ \bm {\hat{q}}} =\bm \mu_{\bm q| \bm x^{\text{target}}}$.
This process is repeated online during a roll-out and only requires the integration of \eqref{eq:coupled_osc_learning} and the inference in \eqref{eq:inference}.
By the end of the roll-out, the cost of using this particular weight vector is then assessed.
Once the optimization is finished, the weights are fixed and a roll-out can be run repeatedly in single-stroke cases or indefinitely in cyclic cases.

\section{Results}

This section describes experimental results on a cyclic basketball pushing task using a real human-sized upper body bi-manual humanoid comprised of 17 DoFs (seven DoFs in each arm, and three DoFs on the waist).
While the ball pushing is a cyclic task, we later show experimental evidence on the generality of the method motivated by a single stroke task of handovers.
In the ball pushing task, the ball adds three DoFs to the task as Cartesian positions such that the PPMP encodes 20 DoFs.

\subsection{Phase Portrait Movement Primitives on a Cyclic Ball Pushing Task} \label{sec:expballpushing}

Motivated by the ball passing game described in the introduction, we describe experimental results where a Phase Portrait Movement Primitive (PPMP) was learned to reproduce the skill of dynamically receiving and passing a ball.
Since the robot has a time response slower than the fast movements of the ball, this experiment allows us to validate the anticipative action that was motivated in Figure \ref{fig:introa}.
Also, because the ball can be easily manipulated by an external agent, it is easy to introduce large disturbances into the system to evaluate its robustness.
The ball was attached to a 1.5-meter string hanging from the ceiling such that it would sit in front of the robot when resting.
The string limited the ball's travel range and ensured its return, facilitating the execution of roll-outs during training.
The robot task was to persistently maintain the ball on a limit cycle.
To do so, the robot had to repeatedly push the ball as if it was passing it to someone in front of it, and to smoothly decelerate the ball to avoid bouncing during its return while also preparing for the next push.

\subsubsection{PPMP Design from Motion Capture}

\begin{figure}
\centering
\vspace{-0.2cm}
\includegraphics[width=0.99\linewidth]{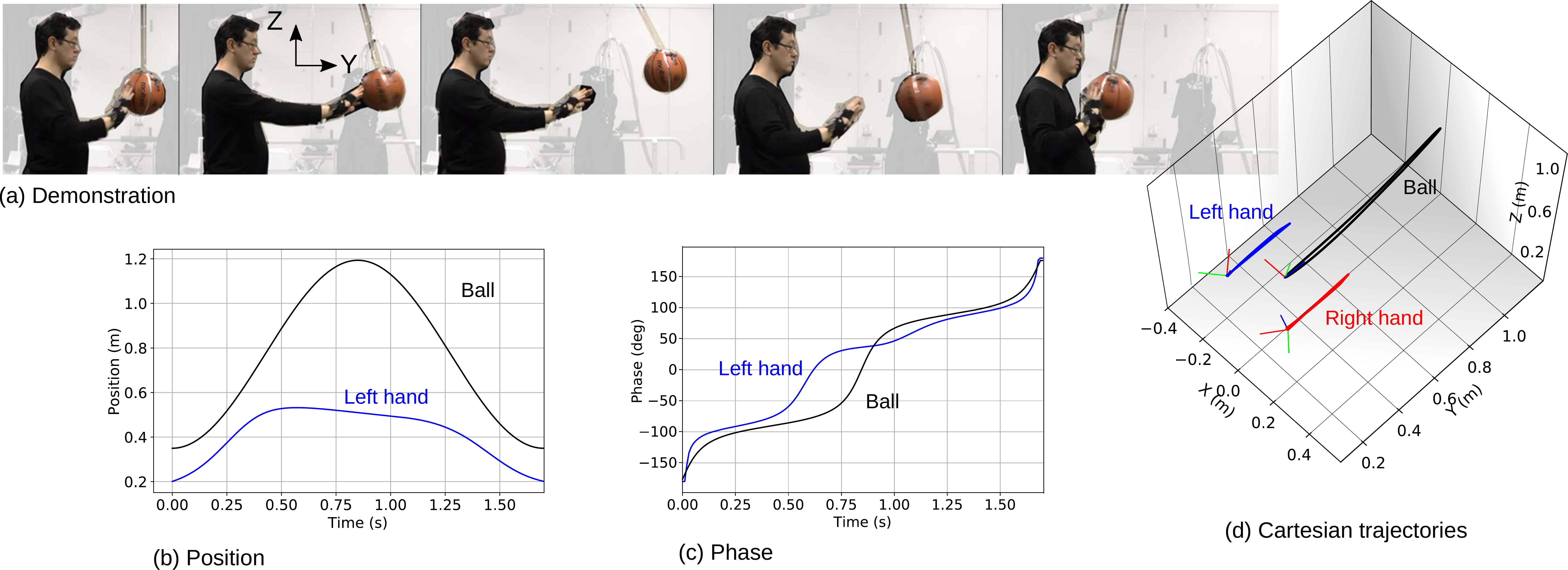}
\caption{
	\textbf{Demonstration of a ball pushing.}
	(a) Demonstration of a ball push-receive task and the resulting Cartesian trajectories (right). Note that the ball hangs from the ceiling by a rope. The trajectories along the Y direction (b) and their phases (c).
}
\vspace{-0.1cm}
\label{fig:ballonchin}
\end{figure}

To design the PPMP, we executed the procedure previously illustrated in Figure \ref{fig:figure1c} where demonstration trajectories of a ball push-receive were recorded as Cartesian trajectories of $T$ time steps for both the hands and the ball as shown by the row of snapshots in Figure \ref{fig:ballonchin}~(a). 
Markers were attached to the right hand of the demonstrator and on the ball.
The right hand trajectory was mirrored across the sagittal plane of the robot to enforce symmetrical dual-arm trajectories such that 
$\bm{x}^{\text{end-eff.}}_t = [(x, y, z, q)_{\text{left}}, (x, y, z, q)_{\text{right}}]^\intercal \in \mathbb{R}^{14}$ ($q \R{4}$ is the quaternion).
The plot in Figure \ref{fig:ballonchin}~(b) shows the recorded trajectories in the Y direction of the left hand and the ball during a full cycle of pushing and receiving.
By comparing the amplitudes of the movements, it is noticeable that the ball had a much larger travel range than the hands meaning that most of the time, the task is underactuated as ball and hands are not in contact.

The PPMP was designed using inverse kinematics (IK) to find the corresponding joint angle trajectories of the robot based on the recorded human's hand trajectories
$\bm q^{\text{robot}}_{1:T} = \text{IK}(\bm x^{\text{demo}}_{1:T})$.
These joint trajectories were paired with the trajectory of the target ball $(\bm q^{\text{robot}}, \bm x^{\text{ball}})_{1:T}$.
To avoid multiple demonstrations we artificially generated perturbed simulations using the real demonstration as a nominal trajectory, a procedure that is explained in detail in the Appendix.
Figure \ref{fig:ballonchin}~(c) shows the corresponding demonstrated phases of the agent and the ball, both computed with \eqref{eq:phaserobot}.
Note that we used the Y coordinates of the movement as it is the direction that measures the distance between the ball and the robot, and thus describes the phase of the interaction.

\subsubsection{Policy Search on the Real Robot}

\begin{figure}
\centering
\vspace{-0.2cm}
\includegraphics[width=0.8\linewidth]{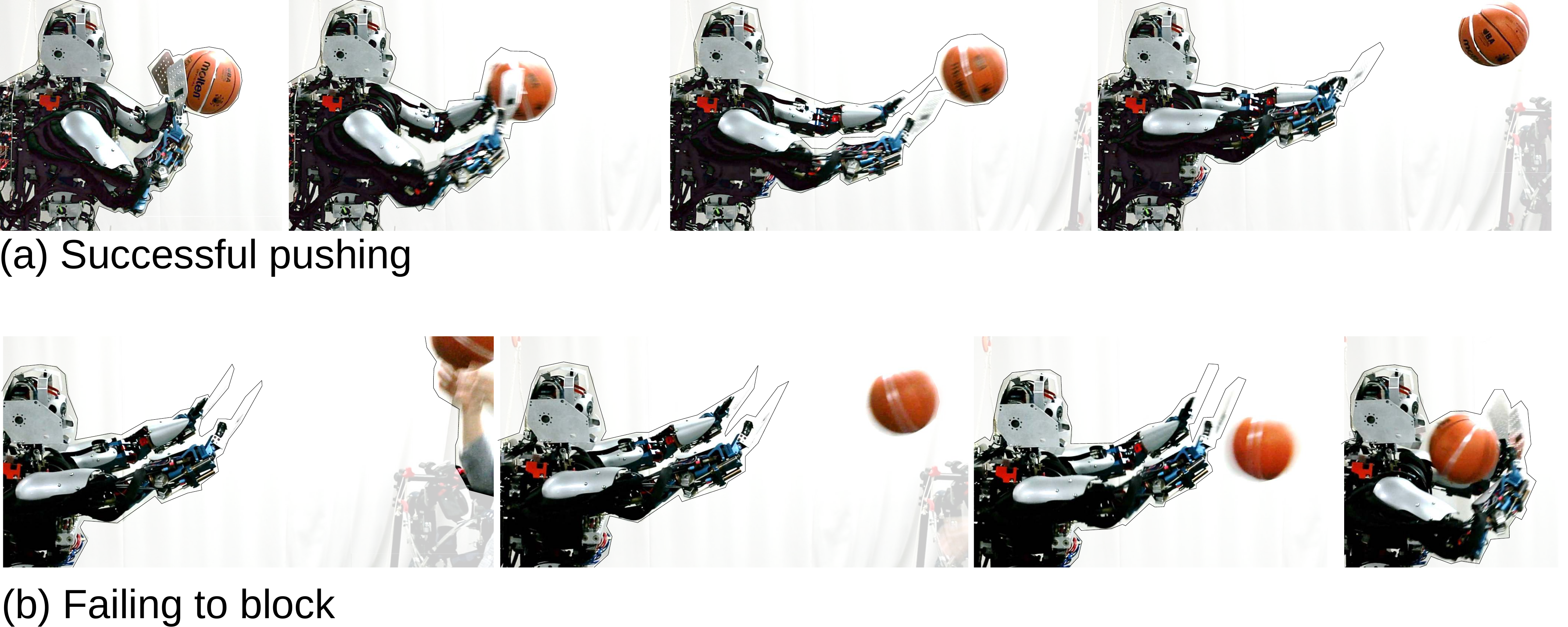}
\caption{
	\textbf{Full cycle with linear coupling.}
	Using constant coupling parameters $K$ and $\alpha$ the robot could push the ball but could not block it in time during its return indicating the insufficiency of the phase predictor to cover different regimes of the cycle.
}
\vspace{-0.1cm}
\label{fig:ballonchin2}
\end{figure}

Our initial attempts in using constant coupling parameters on the phase predictor were unsuccessful, an indication that a nonlinear parametrization is necessary for this task.
As shown in Figure \ref{fig:ballonchin2}, while the dynamics of the coupled oscillators could be tuned to properly push the ball (upper row), the same settings did not succeed when receiving the ball (bottom row). 
As results show, to forcefully push the ball away, the robot must aggressively track the ball phase with a positive phase shift which sets its phase to be ahead of the ball.
Conversely, to softly decelerate the incoming ball and avoid bouncing, a smooth and compliant phase tracking where the robot starts largely advanced in phase but recedes as the ball approaches is necessary.
It is important to keep in mind that although the optimization is only on the parameters of coupled oscillators, as it was shown in Figure \ref{fig:RL}, its effect passes through the kinematics of phase portraits, such that the RL procedure consolidates the full dynamic-kinematic spaces as a single policy.

We follow the procedure of Section \ref{sec:rl} to optimize the policy. 
Since the goal of the task was to maintain the ball on a persistent limit cycle we transcribed this requirement with the cost
\begin{equation}
c_t(y_t) = v_1  \  \sum_{j=1}^{17}( \ddot{q}_{t})^2_j   + v_2 \ | y^{\text{goal}} - y^{\text{ball}}_t| 
+v_3 \ ( 0.5 |y^{\text{left}}_t - y^{\text{ball}}_t| + 0.5|y^{\text{right}}_t - y^{\text{ball}}_t| ).
\label{eq:cost}
\end{equation}
where $\ddot q_t$ is the joint acceleration at time step $t$, $y^{\text{goal}}$  is a goal value set to 3 meters which encouraged the robot to strongly push the ball far away from itself.
The left and right end-effector positions are given by $y^{\text{left}}_t$ and $y^{\text{right}}_t$.
The last term rewards the robot for keeping its end-effectors and the ball distant from each other, which is only possible if the robot learns to constantly push the ball away from itself.
{\myrev
The weights of each component $\{v_1, v_2, v_3\} = \{10, 5, -20 \}$ were manually designed as in the case with reinforcement learning or optimal control studies.
We set the weights, run a few updates, and observe if the cost decrease is aligned with strong ball pushes and smooth movements.
For example, a decrease in cost during the RL updates leading to jerky robot movements indicates that the value of $v_1$ is too small relative to the other parameters. 
}
Since the task is cyclic, a roll-out was defined by its duration and its cost $C$ was computed as the average of the instantaneous costs
\begin{equation}
C( y_{1:M}) = \sum_{t=1}^M  c_t (y_t) /M
\label{eq:totalCost}
\end{equation}
where $M$ is the total number of time steps during the roll-out. Each roll-out was set to last 30 seconds allowing the robot to attempt 15 to 17 pushes per trial.

\begin{figure}    
\centering
\includegraphics[width=1\linewidth]{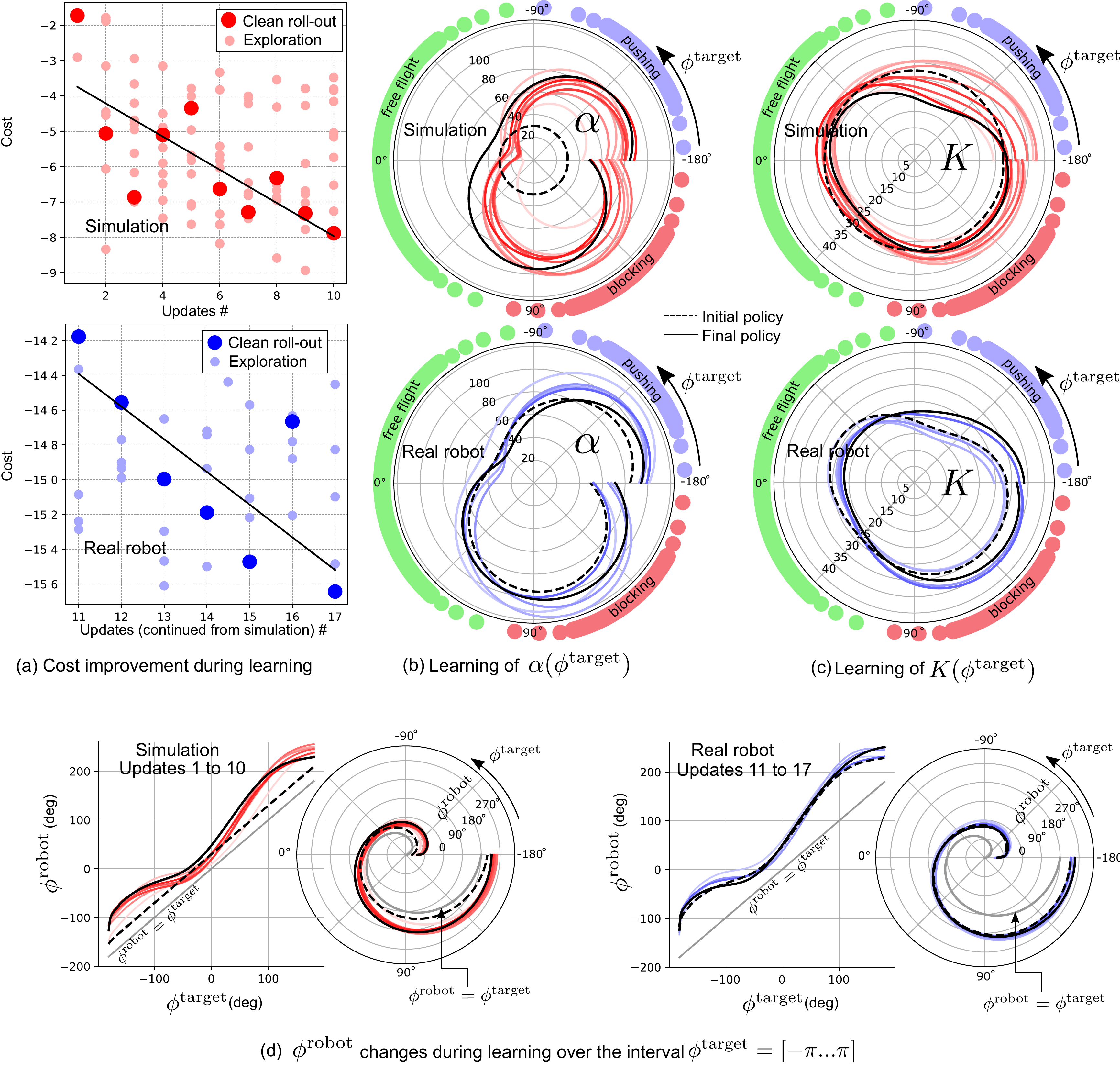}
\vspace{-.5cm}
\caption{
	\textbf{Cost and policy improvements on the repetitive ball pushing task}. 
	{\myrev
		(a) Cost decrease in simulation (top) followed by optimization in the real robot (bottom).
		The big and small circles represent the cost of exploitation (clean) and exploration (noisy) roll-outs, respectively. 
		The lines are first-order polynomials fit to the clean roll-outs.
		Although it is not possible to make strong guarantees on the convergence (e.g. monotonic), the trend lines show clear improvement within the set of updates.
		The optimization of $\alpha(\phitarget)$ (b) and for $K(\phitarget)$ (c) according to the updates in simulation (top) followed by refinement using the real robot (bottom).
		The dashed curve shows the trajectory of the initial policy as a function of the phase of the ball $\phitarget$, and the solid black curve the final optimized policy.
		(d) The phase of the robot over a period of $\phitarget$ at each policy update. The real robot case is a continuation of the policy optimized in simulation. 
		As a reference, the thick gray curves show the case where the phase of the robot is identical to the target (i.e. $\phi^{\text{robot}}=\phitarget$).
	} %myrev
}
\label{fig:policy_search}
%
% code for rebuttal
%    \label{fig:policy_search}
%    Code to genereate figures are:
%    ./Dropbox/ATR/TECHNICAL/2020-03-14_ball_pushing_NN_PolicySearch/B_phase_experiment/main_rebuttal_NN_polar_policy_search.py
\end{figure}

To allow for fast and aggressive initial explorations, the first 10 policy updates---each update consisting of 10 roll-outs---were run in a simulated environment under large exploration noise.
Subsequently, seven additional policy updates---each consisting of five roll-outs---were run on the real setup totaling 17 minutes of training.
The top plot in Figure \ref{fig:policy_search} (a) shows the reduction of cost in simulation.
The bottom plot shows the continuation of the training using the real robot.
At each update cycle, the larger circles (red in simulation and blue in real experiments) represent the cost of a ``clean'' roll-out, that is, the roll-out where the best current policy was evaluated.
The small light circles represent the cost of ``exploration'' roll-outs, that is, the roll-outs where the base policy was added with additional noise in an attempt to generate better (lower cost) versions of the base policy.
The variance of costs was larger in simulation than in the real setup due to the larger exploration of the parameters.
Compared to the costs in simulation (which aggressively decreased from $-2$ to $-9$ due to larger exploration noise) the absolute cost on the real robot was lower (decreasing from $-14$ to $-16$) as the ball had a larger travel range on the real setup, thus, generating roll-outs of higher rewards. 

{\myrev
Figure \ref{fig:policy_search} (b) and (c) show the changes in the policy in simulation (top plot with curves in red) and subsequently after switching to the real system (bottom plot with curves in blue).
The final solutions in simulation and on the real robot are shown as the solid black curves.
Note that because the optimization in the real robot is a continuation of the learning in simulation, the initial policy of the former is the final policy of the latter.
From these phase plots, a qualitative observation from the real robot phase plots shows that the coupling functions have larger values within the range of the pushing and blocking regimes.
The lower values, particularly evident on the bottom plot (b), occur during the free flight regime covering mainly the second and third quadrants when the robot cannot actuate the ball.
The pushing starts at the first quadrant, requiring a large phase difference $\alpha$ to put the motion of the robot ahead of the ball. 
The high stiffness $K$ also enforces a close phase tracking at this critical stage.
As the ball returns to the robot on the fourth quadrant, the phase difference increases again peaking at around 90$^0$ such that the hands returned faster than the ball, and the difference decreases at the end (180$^0$), an indication that the robot attempts to approximate its phase with the ball phase for a synchronized deceleration under strict phase tracking (large $K$ values).

Note from (b-c) that although the policies refined during the real experiments did not drastically change the profile of the curves initially learned in simulation, it is noticeable that the radii of the paths during real experiments tended to increase.
This increase suggests compensation for the dynamics of the real robot, which presents delays, tracking error, and compliance; while in the simulator the robot was as an ideal reference tracker with infinite bandwidth.

Finally, Figure \ref{fig:policy_search} (d) shows the changes in the trajectories of the function  $\phi^{\text{robot}}(\phitarget)$ as $\phitarget$ covers the range $[-\pi, \: \pi ]$ at each policy update.
The curves are computed by integrating \eqref{eq:coupled_osc} using the respective values of  $\alpha(\phitarget)$ and $K(\phitarget)$ as previously shown in (b-c).
As a reference, the thick gray curves represent the theoretical case where $\phi^{\text{robot}}=\phitarget$.
It is evident that the phase of the robot is usually advanced with respect to the target, with the lowest values occurring close to 0$^0$ which refers to the middle of the free-flight regime.
Note that because $\phitarget$ is never constant (unless the ball is stopped) the phase error ($\phitarget-\phi^{\text{robot}}$) is not expected and does not necessarily have to vanish for the task to be successful.
}

\subsubsection{Evaluating the Optimal Policy with Spatio-Temporal Disturbances}

\begin{figure}
\centering
\includegraphics[width=1\linewidth]{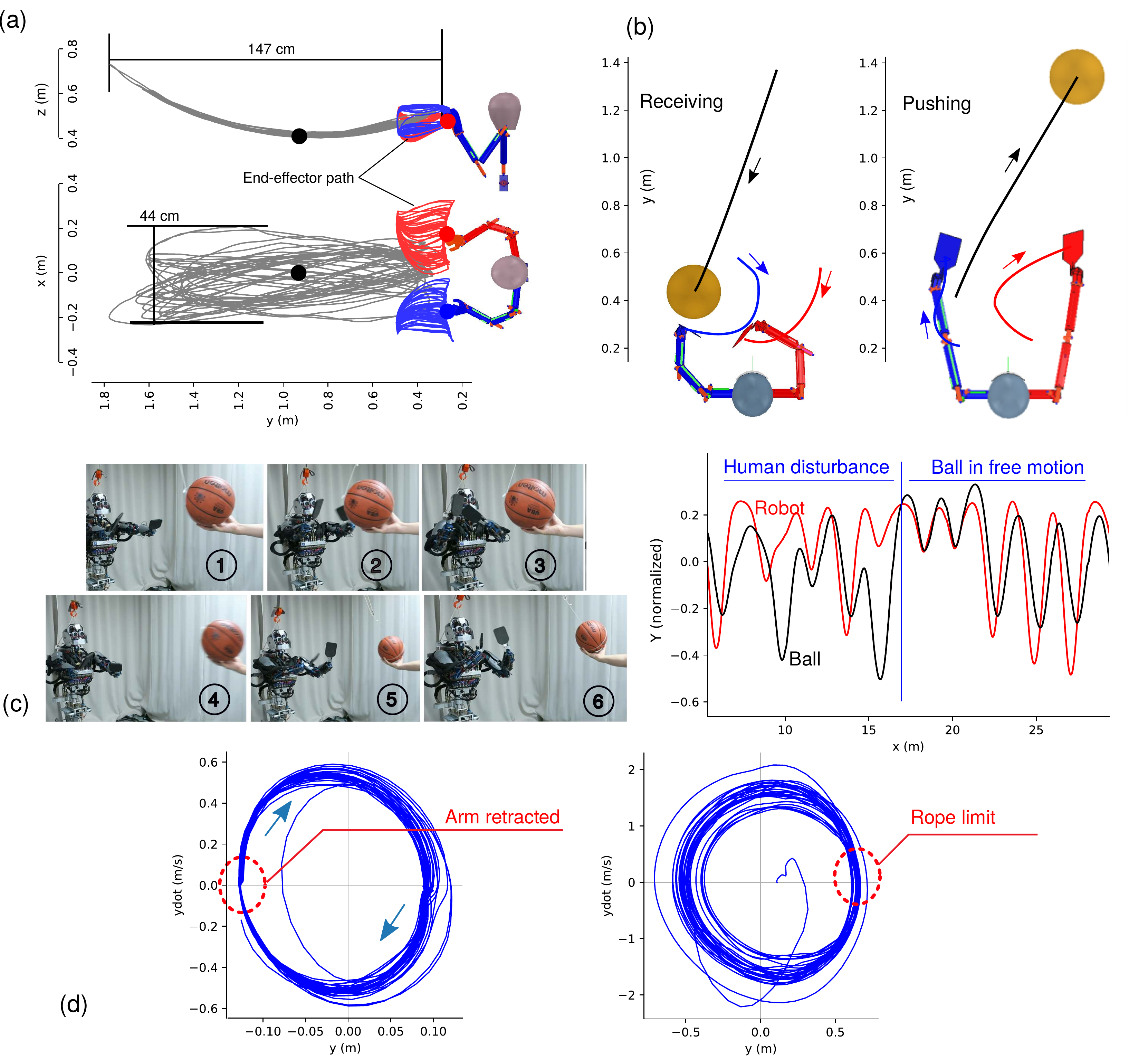}
\caption{\textbf{Extended experiments using the final policy.}
	(a) The paths of the ball and hands as the robot interacted with the ball for one minute.
	(b) Two real trajectories of the ball receiving and pushing the ball. Note the non-trivial asymmetry of hand trajectories.
	The robot's illustration is shown to facilitate interpretation.
	(c) A one-minute trial where a person grabbed the ball and pretended to pass the ball but reversed the trajectory a few times. Note from the snapshots that the robot tried to adapt by changing its pose.
	The curves show the left-hand trajectory in the Y direction as the robot responded to the changes in the ball flight. 
	The phase plane trajectories of an undisturbed experiment are shown in (d). The highlighted areas represent the physical constraints of the robot joint limits and the length of the rope.
	A video of the ball pushing experiment is provided with this article and can also be accessed via the link
	\url{https://youtu.be/IgC-n-HrAWI}.
}
\label{fig:expresults}
\end{figure}

Figure \ref{fig:expresults}~(a-b) shows the use of the optimized policy on undisturbed ball trajectories as the robot repeatedly pushed and received the ball for an entire minute.
While the plotted trajectories were obtained in real experiments, the figures are overlaid with a graphical image of the robot to facilitate interpretation.
As shown in the side-view in (a), the strongest pushes moved the ball almost $1.5$ meters away from the robot in the Y direction.
Although the ball was not externally disturbed, the lateral swing of the ball barely stayed on the sagittal plane of the robot (the plane that cuts the body symmetrically in right and left sides) creating elliptical paths $44$ cm wide and forcing the robot hands to move sideways (see subplot (a)).
The large variation of ball trajectories is the result of uneven dual-arm pushes, in part, due to the difficulties in setting a perfectly symmetric scenario.
Subplot (b) shows two segments where the robot moves to block the ball (left), and subsequently pushes it (right).
The left and right hand's non-trivial and dissimilar paths act in concert to achieve a successful manipulation of the ball.
These complex actions are the result of imitating the coordination implicit in the human demonstration.

We validated the fast adaptive mechanism of PPMPs employing spatio-temporal disturbances.
The robot reacted almost immediately---with a bandwidth between 30-60 Hz which was limited by the frequency of the RGB-D camera.
Figure \ref{fig:expresults}~(c) shows a trial where a person often disturbed the task.
The sequence of snapshots shows the moment someone grasped the ball mid-flight and pretended a pass to the right and then to the left of the robot. 
Note from frames 1 and 4 that the arms of the robot are fully extended as the ball is moving away from the robot.
In frames 3 and 6, the robot preemptively positioned its hands to receive the ball at the appropriate positions as the phase of the robot was largely advanced with respect to the ball.
This behavior is the exact one described in our initial motivation in Figure \ref{fig:introa} where the robot advances its phase to wait for the ball landing.
The normalized curves at the right of Figure \ref{fig:expresults}~(c)  show the trajectories of the ball and the robot's hand moving along the pushing direction.
In the first segment, the ball flight was disturbed by someone vigorously rocking the ball back and forth.
In the second half, the ball was released and the robot could graciously recover and bring the ball back to a limit cycle (refer to the accompanying video, which is also accessible here \url{https://youtu.be/IgC-n-HrAWI}, to better understand the dynamics and intensity of disturbances applied during the experiments).

Figure \ref{fig:expresults}~(d) shows the phase plane trajectories corresponding to an undisturbed limit cycle of the ball.
The low variance at the leftmost part of the phase of the robot is due to the arm being in a fully retracted pose, ready for a push.
In the case of the ball phase plane, the variance is lower at the rightmost part of trajectories as the ball was achieving its maximum travel range constrained by the length of the rope.
The phase-plane trajectories of the ball evidence a stable limit cycle.
In extended experiments, the robot could maintain the ball on a limit cycle for more than five minutes, which resulted in more than 120 consecutive pushes.

This experiment allowed us to evaluate the responsiveness of the PPMPs when adapting to the extremely large and arbitrary spatio-temporal disturbances introduced by a human manipulating the ball.
Also, the experiments provided empirical evidence that the dimensionality of the PPMPs makes it feasible to train the policy with reinforcement learning and to run the algorithm at 30 Hz on a system comprised of 20 DoFs in total (17 DoF robot of the robot and three DoFs of the moving ball).
It is worth noting that the PPMP was learned in a semi-model-free setting, where the only modeling assumption was the two coupled oscillators.
In contrast, in online optimal control, the designer is first met with the challenge of modeling, identifying parameters, and validating the predictions of the bi-manual contact forces which, per se, is arguably more challenging (if not impractical depending on the required accuracy) than the entire design of the PPMP itself.
Only after the system identification step is overcome, the designer would be able to proceed with the implementation of the necessary optimization routines.

\subsection{PPMPs on Single Stroke Tasks: Handover Case}

PPMPs are not exclusive to cyclic tasks.
The methodology can be applied without modifications to single-stroke tasks as well. 
As in the cyclic case, the only mild assumption is that the dynamics of the coupled oscillators are sufficient to describe the temporal interactions of the task at hand.
Experiments in handovers were chosen not only because handovers are one of the most studied tasks in the field of physical human-robot interaction \cite{strabalaSeamlessHumanrobotHandovers2013} but also because it is a task where \emph{timing} is very relevant for fluid and natural interactions. 
{\myrev
Note that while in the single-stroke PPMP case, the phase range of the oscillator is constrained to lie only on the first two quadrants such that there is no path of return, other researchers have achieved a continuous discrete-to-oscillatory behavior by modulating the parameters of the oscillator itself \cite{jouaitiCPGbasedControllersCan2018}, 
}

\subsubsection{Handover Demonstrations}

\begin{figure}    
\centering
\includegraphics[width=1\linewidth]{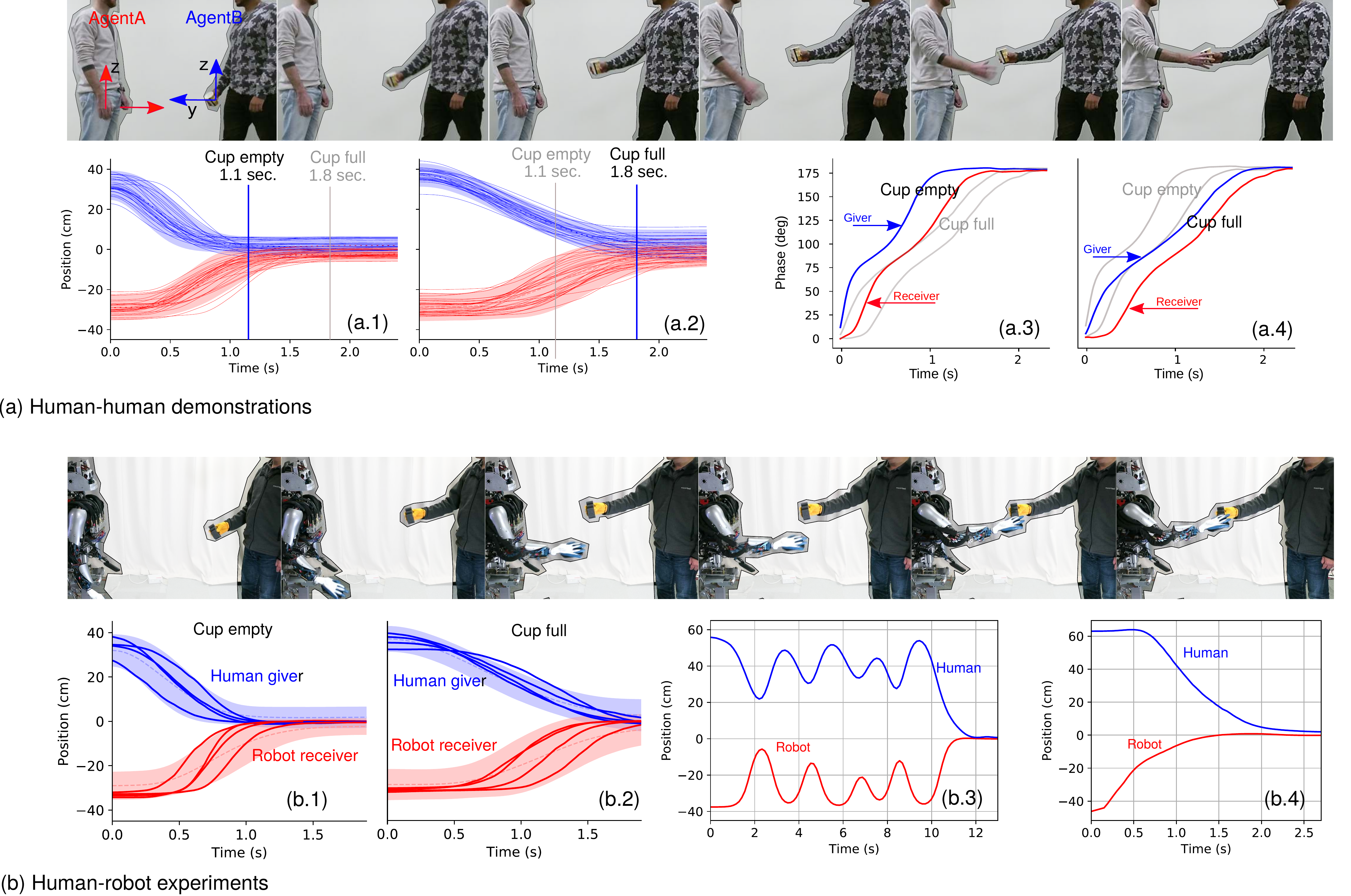}
\vspace{-.5cm}
\caption{\textbf{PPMP in discrete handover tasks}. 
	(a) Snapshots show one instance of human demonstrations using markerless skeleton tracking.
	(a.1, 2) Show 30 demonstrations for the cases where the cup is empty and filled with water, respectively.
	(a.3, 4) The progress of the phases using the averaged values on the empty and full cup cases. The grey curves represent the opposite cup state to facilitate comparison.
	(b) Experiments using the real robot where agent A was replaced by the humanoid. 
	(b.1,2) Examples of handovers using the real robot as a receiver.
	Subplot (b.3) shows the case where the human tricks the robot by pretending passes but retracting his hand a few times.
	(b.4) The robot acting as a giver where its phase evolves in open-loop.
	A video of the handover experiment is provided with this article and can also be accessed via the link
	\url{https://youtu.be/iAlu5ZH0SSM}.
}
\label{fig:handover_demo}

%    
%    (Settling time metrics (5 percent of final state)
%    Setting time for metrics 5 percent
%    Agent1 0.900, agent2 1.409 (sec)
%    
%    
%    (Settling time metrics (5 percent of final state)
%    Setting time for metrics 5 percent
%    Agent1 1.647, agent2 1.984 (sec)
%    

% Code for demonstrations
%     /localgjm/Dropbox/Papers/my/GJM/current/ScienceRoboticsSpecialIssue/TeX_submission_1st_November/figures_handover/python_code_figures/main_demonstrations.py
% Code for plotting expermental results
%     /localgjm/Dropbox/Papers/my/GJM/current/ScienceRoboticsSpecialIssue/TeX_submission_1st_November/figures_handover/python_code_figures/main_experiment.py
% Code for plotting simulated results
%     /localgjm/Dropbox/Papers/my/GJM/current/ScienceRoboticsSpecialIssue/TeX_submission_1st_November/figures_handover/python_code_figures/main_simulation.py
%
%                % OLD, probably deprecated
% human-human plots
%     /localgjm/Dropbox/ATR/TECHNICAL/2018-08-06_HRI_real_robot/code_phase_from_human_observation/main_get_phase_and_trajectory.py

% human-robot plots    experiments
%    /localgjm/Dropbox/Papers/my/GJM/current/ScienceRoboticsSpecialIssue/TeX_submission_1st_November/figures_handover/experiment/main_plot_for_paper.py
\end{figure}

We implemented a single PPMP to interact with a human partner under different timings in handovers.
To this end, we recorded a total of 30 demonstrations of a cup handover where the cup was empty and 30 demonstrations where the cup was filled with water.
The upper row in Figure \ref{fig:handover_demo}~(a) shows a sequence of snapshots of one demonstration instance.
The right hand trajectories of each demonstrator were recorded as a sequence of Cartesian coordinates $\bm x^{\text A}_{1:T}, \bm x^{\text B}_{1:T}$, with 
$\bm x^{}_t = [x^{}, y^{}, z^{}]^\intercal \in \mathbb{R}^{3}$ via motion capture.
The first snapshot shows the convention of coordinate frames where the horizontal and vertical axes are on the sagittal plane of the agent.
Figures~\ref{fig:handover_demo}~(a.1, a.2) show the two sets of demonstrations as a distribution (mean \mbox{$\pm$ two} standard deviations) for the case when the cup was empty and full, respectively.
The figures also indicate the mean settling time of the handover.
As expected, when the cup is empty the settling time is shorter than when the cup is full (see indications on plots (a.1, a.2)).
In the context of phase dynamics control, the underlying hypothesis is that the giver acts as the phase reference, and the receiver adapts its phase according to the giver's progress.

Subplots (a.3-4) summarizes the phase relationship by the average progress of the human phases for both empty and full cup cases.
The phase of each agent was computed with \eqref{eq:phaserobot}.
To facilitate comparison, the figure shows the opposite cup condition (full/empty) in grey.
It is noticeable that when the cup is empty, the phase of the slower agent (the receiver) reaches $180^0$ at around 1.5 seconds, while when the cup is filled with water its phase achieves the same value at around 2 seconds.
In this single stroke case, only the first and second quadrants of the phase plane are traversed as no returning path exists.

\subsubsection{Handover Experiments with the Real Robot}

Compared to the cyclic ball pushing, controlling the handover is possible with  constant coupling parameters  $K$ and $\alpha$.
In this case, it is not hard to tune these two values by hand.
The plots in Figure \ref{fig:handover_demo}~(b.1) show the trajectories of four different experiments where the trajectories of the human and the robot hands movement are overlaid on the distribution of demonstrations corresponding to the empty cup case.
The values of the coupling were set to $K=30$ and $\alpha=-65^0(\pi / 180^0)$.
Figure \ref{fig:handover_demo}~(b.2) show four similar cases for slower movements of the giver handing over a cup full of water.
Compared to the empty cup response, the more sluggish response of the robot was achieved by decreasing the value of the coupling stiffness to $K=20$.

The humanoid upper body was used to replace the role of agent A as the receiver of the object.
At each updated position of the giver's hand, estimated using the color of the glove and a depth sensor, the PPMP provided the vector of joint angles $\bm q_t$ that defined the corresponding pose of the robot.
The computed joint angles were used as reference angles for the low-level position tracking controller of the robot.
Figure \ref{fig:handover_demo}~(b) shows a sequence of snapshots of a handover where the robot receives the object from the human giver.
All experiments used the same phase portrait with different coupled oscillator parameters.

The fast phase adaptation and prediction of PPMP make the robot react as if it had a sense of time similar to humans.
This feature is shown in Figure \ref{fig:handover_demo}~(b.3) where during the handover the human purposely rocked his hand back and forth before finally handing over the object.
The robot could adaptively advance and retract its hand, in the same way, people would do in such a situation.
To better understand the dynamics of the experiments, an extensive sequence of handovers can be watched in the accompanying video.

For completeness, and to demonstrate the flexibility of the oscillator dynamics formulation, we swapped the roles between the human and the robot by running the phase estimator in open-loop.
That is, the indexes of the phase-portrait $P(\bm x, \bm q)_{\phi^{\text{robot}}}$  were run as a function of time  where 
timing was reproduced from one of the demonstrations, $\phi^{\text{robot}}_{1:T} = \phi^\text A_{1:T}$.
In this way, the robot moved independently of the human partner's progress, although the position of its hands was still being coordinated with that of the human.
One particular instance is shown in Figure \ref{fig:handover_demo}~(b.4).

%\subsubsection{PPMP vs Hard-Coded Trajectories}

As discussed, one of the main advantages of PPMPs is to be a semi-model-free approach which only requires the general dynamics of coupled oscillators for phase prediction.
In contrast, take for example the recent work of Pan et al. \cite{Pan2019IROSfasthandover} where a handover controller was implemented using 
B\'ezier curves as joint trajectories of a 7-DoF robotic arm.
To account for the many stages of a $  $handover, Pan et al. used a hybrid automaton approach to activate the corresponding actions.
PPMPs can potentially simplify the number of high-level rules or discrete states in such automaton as the coupled oscillators account for most of the motion adaptation, including the robot's arm retraction when the human decides to stop the handover midway.
Also, PPMPs allows for dynamical modulation of the timing along the entire trajectory while in \cite{Pan2019IROSfasthandover} only the initial delay could be manually adjusted.
The reader is invited to watch the video \url{https://www.youtube.com/watch?v=w1Ff4nqcUvk} of Pan et al., and the accompanying handover video of this paper \url{https://youtu.be/iAlu5ZH0SSM}.
Qualitatively, both methods seem to generate similar robot reactions. 
However, while in \cite{Pan2019IROSfasthandover} the robot acted based on pre-designed B\'ezier curve trajectories and automaton rules, PPMP actions were obtained by using imitation and the oscillator model which involved the adjustment of only two parameters.
%to pave the way for autonomous learning.

%As discussed, one of the main advantages of PPMPs is to be a semi-model-free approach which only requires the general dynamics of coupled oscillators to generate predictive adaptation.
%In contrast, take for example the recent work of Pan et al. \cite{Pan2019IROSfasthandover} who implemented a handover controller where the joint trajectories were hard-coded for each degree of freedom of a 7-DoF arm.
%To account for the case where the robot needed to retract its arm, Pan et al. used a kind of finite-state machine approach to activate the action for arm retraction.
%In PPMPs, the arm retraction is achieved without additional high-level rules, as it is simply the result of the adaptation provided by the coupled oscillators.
%Also, PPMPs dynamically control the timing along the entire trajectory while in \cite{Pan2019IROSfasthandover} only the initial delay can be manually adjusted.
%The reader is invited to watch the video \url{https://www.youtube.com/watch?v=w1Ff4nqcUvk} of Pan et al., and compare with the accompanying handover video of this paper \url{https://youtu.be/iAlu5ZH0SSM} while qualitatively observing the similarities in the robot's reaction.
%While in \cite{Pan2019IROSfasthandover} a robot acting based on engineered trajectories and human-made rules allowed the authors to focus on the effect of timing in social human-robot interaction, our focus with PPMPs is to eliminate hard-coding or engineering of tasks to pave the way for autonomous learning.

\section{Discussion} \label{sec:discussion}

This section discusses aspects of PPMPs concerning other primitive representations.
In particular, we apply the method in the cyclic task of walking to compare it against a periodic Dynamical Movement Primitive used as a Central Pattern Generator.
{\myrev We also present some other observations related to direct feedback tracking, the temporal smoothness of the PPMP, and the limitations on the scope of a PPMP}

\subsection{Phases in other Movement Primitive Representations}

In manipulation, many authors have proposed different ways to provide primitives with time-independence.
These can vary from the use of Hidden Semi-Markov Models to learn the transition dynamics of the movement\cite{calinon2011encoding}, to scaling the velocity of a learned dynamics used for trajectory prediction \cite{kim2010learning}, to using the ratio between the current robot state and the remaining path till the goal \cite{englert2014reactive}, to cite a few (a concise review with many approaches can be found in \cite{kim2014catching}).
Dynamical Movement Primitives (DMPs) \cite{ijspeert2013dynamical} have long suggested the explicit use of phases to replace time, and more recently, Probabilistic Movement Primitives (ProMPs) \cite{Paraschos2013NIPS} also followed the same idea.
However, in real robots, the use of phases in existing formulations has been quite simplistic; mainly as an open-loop signal to synchronize multi-DoF systems and to adapt the speed of movements.
Here, we state three advantages of PPMPs concerning existing movement primitive formulations.

\textbf{Fast Predictions in Phase-Space}. PPMP provides a principled and efficient way to control phases.
Also, note that its oscillator can be used with existing movement primitives explicitly parameterized by phases such as DMPs and ProMPs.
Compare the phase mechanism of PPMP, DMP, and ProMP cases, 
\begin{equation}
\begin{split}
\dot{\phi}^{\text{robot}}       & = \omega + K \sin (  \phi^{\text{target}} - \phi^{\text{robot}} + \alpha )   \text{ \ \ \   PPMP, }\\
\tau \dot{ \phi}^{\text{robot}} & = -\alpha_{s} \phi^{\text{robot}}  \text{\ \  \ \ \   DMP in the discrete case, }\\
\tau \dot \phi ^{\text{robot}}  & = 1                                \text{\quad \quad  \quad \  \ \ \ \ \  DMP in the cyclic case,}\\
\phi^{\text{robot}}             & = f(t)                             \text{\quad \quad   \ \   \ \ \  ProMP,}    
\label{eq:phase_dmp}
\end{split}
\end{equation}
where $\alpha_s$ is a positive constant value, $\tau$ is the system time constant, and $f(t)$ is any monotonically increasing function.
Since the phase of DMPs and ProMPs evolve in open-loop, temporal adaptation relies on external mechanisms to correct the phase (e.g. by using Kalman filter when model parameterization is possible \cite{kober2010movement}, or by learning models for unknown object dynamics as it was done in \cite{kim2014catching}).
In contrast, Phase Portrait Movement Primitives (PPMPs) exploits the use of the coupled oscillators for predicting phases, thus achieving much faster adaptation under lower computation and simpler modeling assumptions.
%On the other hand, the reactive feedback nature of the coupled oscillators should not be confused with the inability to predict states in the future as a positive phase shift advances the robot ahead of the current temporal evolution of the interaction.
%

\textbf{Scalability for Joint-Space Control}.
A significant computational advantage over DMPs for joint space control is the fact that a single PPMP is used regardless of the number of degrees-of-freedom of the robot. 
This is possible because the same PPMP encodes the correlation of all degrees-of-freedom of the robot, which is a feature also provided by ProMPs and GMMs.
On the other hand, for joint-space DMPs and joint-space GPs \cite{schneiderRobotLearningDemonstration2010, maedaActiveIncrementalLearning2017}, the number of primitives scales with the number of joints.

\begin{figure}
\centering
\includegraphics[width=0.98\linewidth]{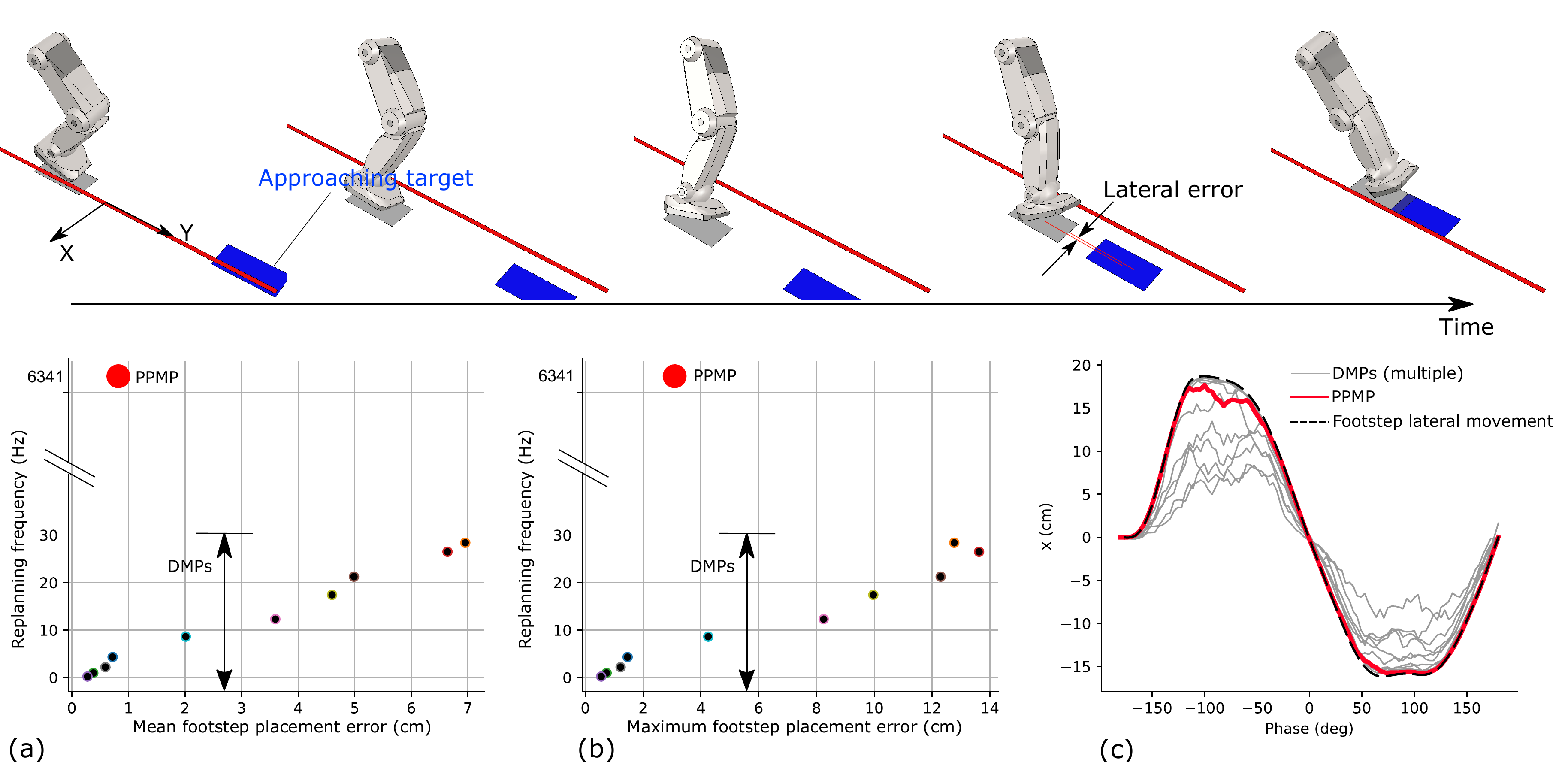}
\caption{
	{\myrev
		\textbf{DMP vs PPMP in a rhythmic task with a simulated robot.}
	}        
	Upper row: a sequence of snapshots where PPMP is used to infer the joint angles of a {\myrev simulated} robotic leg given the footstep placement while executing a walking cycle.
	The phase oscillator dynamics define the progress of the walking pattern while the lateral target motion dictates the foot placement. 
	(a-b) when using DMPs, smaller foot placement errors (mean and maximum) can be obtained by allowing the optimizer to converge. In turn, waiting for convergence leads to low re-planning frequencies.
	By controlling the maximum number of allowed iterations, the re-planning frequency can be increased up to 30 Hz, at the expense that the error of the DMP also increases as the optimization process is truncated.
	In (c) the effect of the planning frequency is observed in terms of the DMP trajectories. Trajectories with larger errors were planned faster. In general, PPMP plans at kHz order while achieving the same error of DMPs planned at 10 Hz.}
\label{fig:walking}
\end{figure}

\textbf{Fast Spatial Adaptation on Cyclic Tasks}. 
The third advantage of our method is evident in cyclic tasks.
PPMPs natively allow the robot pose to be adjusted instantaneously, at each time step.
Periodic DMPs require optimization over the entire limit cycle as its goal attractor can only modify the averaged behavior of the cycle but not the instantaneous position at each time step.
{\myrev
We illustrate this difference by a simulated robotic walking task as shown in Figure \ref{fig:walking} where PPMPs and DMPs are compared quantitatively.
}
As shown by the snapshots, in this task, the incoming desired footstep placement (assumed given by a perception system) moves sideways and the robot must adapt the foot position laterally while the walking cycle evolves. 
On rhythmic DMPs, this adaptation requires optimizing the forcing function parameters with a cost that penalizes for lateral error position
In PPMPs adaptation requires solving $\bm q \sim P(\bm q | \bm x^{\text{target}})$ which has a closed-form solution for Gaussian distributions.
Figure \ref{fig:walking}~(a)  show that the DMP error in footstep placement increases with the increase of the replanning frequency\footnote{As an optimizer, we used the Pi$^{\text{BB}}$ \cite{stulp2012policy} algorithm and controlled the replanning frequency by changing the maximum number of parameter updates.}.
In the case of PPMP, no optimization is required and the frequency of replanning runs two orders of magnitude faster\footnote{Both methods were implemented in Python and run on the same computer.
}.
Finely optimized DMPs can achieve less error than the PPMP at the expense of slower updates (less than 5 Hz).
This is because PPMP computes the robot pose by inference. 
As such, its accuracy depends on how close the true distribution fits the assumption of normally distributed spatial models. 
For fine tasks that demand accuracy, a mixture of PPMPs or optimization on the PPMP solution may be necessary.\\

\subsection{Direct Tracking via Position Feedback}

In general, direct feedback tracking of the position of the ball or the hand of a human (e.g. with a Cartesian controller under visual servoing) cannot accomplish the same level of task complexity of movement primitives, including PPMPs.
In a sense, direct feedback provides robotics reflexes: given a stimulus, it outputs an instantaneous action that does not involve reasoning over future states.
On the other hand, PPMPs make use of the kinematic distributions and positive phase shifts to advance the pose of the robot concerning the current phase of the target.
For example, the motor actions of receiving and pushing the ball are not solvable by pure reflexes given by a Cartesian tracking controller.
Since a feedback controller attempts to decrease the tracking error, the robot would try to move the hands until they touch the ball, but not to push it.
Also, because the ball moves faster than the arm, direct tracking of position is prone to fail.
In contrast, the learned PPMP policy allowed the robot to act in a predictive manner, by positioning the hand at the right location before the ball arrived and by later pushing the ball far away, a feature that no visual servoing controller can provide.

\subsection{Spatio-Temporal Smoothness of PPMPs}

One distinct characteristic of PPMPs is that the joint distributions on the phase portrait are independent of each other.
In principle, the temporal states of the robot are allowed to ``jump'' or ``skip'' in time. 
This is different from most primitive formulations that usually rely on some mechanism to guarantee temporal smoothness to generate suitable robot commands.
In the case of DMPs, the smoothness is due to the use of radial basis functions used to encode the forcing function.
ProMPs similarly achieve this smoothness with the difference that the basis functions encode the positions and velocities.
Other methods where the smoothness is provided by construction due to the appropriate choice of kernels or features are Gaussian Processes \cite{schneiderRobotLearningDemonstration2010} and Gaussian Mixture Models \cite{calinon2011encoding} as primitives.

The mechanism responsible for temporal smoothness in PPMPs is the dynamics of the coupled oscillators, which is governed by a differential equation \eqref{eq:coupled_osc}.
Since the phase is the result of an integral, it can only evolve continuously over time which enforces the temporal smoothness of robot commands.
The kinematic smoothness in PPMPs is a consequence of the robot motion being conditioned on the target motion.
In the experiments of this article, the targets (the ball, the human hand, the footstep placement) moved smoothly in space, that is, they did not ``teleport'', and as a consequence, the conditioned robot movement behaved accordingly.
PPMPs shows promise in future applications under hybrid controllers where hard-switches may occur as PPMPs do not enforce smoothness regarding spatial transitions.

{\myrev
\subsection{Beyond Task-Specific PPMPs}

PPMPs are designed following the definition of \cite{schaalImitationLearningRoute1999}, as an elementary/atomic control policy that is specialized to realize a particular ``goal-directed behavior''. 
Thus, it has not been our primary concern to consider the generality of a single PPMP towards multiple tasks.
Biological systems may be naturally endowed with the ability to generalize/reuse primitives for multiple tasks, but there is no consolidated robot primitive methodology that can reproduce such capability.

A more tractable, usual approach, in robotics, is to leverage previously learned primitives (as policies) and adapt them (e.g. via reinforcement learning) to solve a new task as in transfer learning \cite{taylor2009transfer}.
A related approach to the latter is meta-learning \cite{yuOneshotImitationObserving2018} where the robot leverages large amounts of past experience as a general model, which can be quickly specialized for the task at hand.

Basketball players learn very efficiently to handle the basketball for various (related) tasks: dribbling, throwing and catching the incoming ball in various directions. 
We speculate that for robots also, after learning one of these tasks, we can transfer the learned knowledge significantly using the same primitive representation.
Take the learned skill of repetitive ball pushing presented in Section \ref{sec:expballpushing} where the oscillator parameters were optimized under a pendular movement of the ball as an example.
Using the same primitive to block an incoming ball that is approaching vertically at speeds never observed during the training may lead to poor performance due to differences in dynamics. 
In transfer learning, we could re-train the existing ball pushing skill using reinforcement learning to augment the robot skills for vertical incoming balls.

Large bodies of work have flourished to address the many challenges posed by the use of task-specific primitives as proposed in \cite{schaalImitationLearningRoute1999}.
To cite a few, the challenge of decomposing complex movements into libraries of primitives \cite{niekumIncrementalSemanticallyGrounded2013, lioutikovLearningMovementPrimitive2017}, mixing multiple primitives to generate reaching movements for different targets \cite{udeTaskspecificGeneralizationDiscrete2010}, measuring their coverage and uncertainty to make active requests for demonstrations \cite{maedaActiveIncrementalLearning2017}, assessing their similarities \cite{stark2017distance}, and combining existing primitives into a complex policy \cite{pengMCPLearningComposable2019}.
Thus, although the PPMP formulation here presented is task-specific, the existing body of work aimed at generalizing and transferring primitives are, in principle, applicable.

}

\section{Conclusions}

Real-world implementations of fast humanoid control executing bi-manual manipulation have been 
extremely scarce and existing cases have usually relied on domain knowledge, carefully engineered solutions, and heavy computation.
This article proposed PPMP, a learning control method suited for fast and anticipative tasks with a native capability to estimate temporal dynamics.
Coupled oscillators provide the robot with predictive adaptation to the dynamics of the interaction while the associated joint distributions are used as priors to spatially correlate all degrees-of-freedom in the task.
The only open parameters of the method are the coupling components between the oscillators, rendering a low dimensional representation that is amenable to the use of reinforcement learning in real robots.
While this approach is inspired by observed motor skill characteristics found in animals\cite{lee1980_visuomotor_timetocontact, turvey1990coordination} our main goal is not to reproduce biological systems per se. 
Rather, the PPMP goal is to be a fast humanoid control method for autonomous learning that can be designed with minimal domain knowledge and run under a low computational budget.
In regards to the literature of movement primitives for robot control, building the method from scratch to include a phase predictor led to a method that is not only faster and simpler to implement particularly for rhythmic tasks, but whose scalability is not affected by the number of degrees-of-freedom of the system.
The semi-model-free approach means that PPMPs are predictive in nature without relying on online optimization 
while being efficient enough to have its policy optimized via reinforcement learning on real, high dimensional tasks.

\section*{Acknowledgments}
This article is based on results obtained with the support from the grants JST-Mirai Program No. JPMJMI18B8 Japan, the JSPS KAKENHI No. JP16H06565, and from a project commissioned by the New Energy and Industrial Technology Development Organization (NEDO).

%\section*{Acknowledgments}

\bibliography{Science2NN}

\section*{Appendix} \label{sec:appendix}

\subsection*{Data Augmentation on Ball Pushing Experiments}

The design of the probabilistic phase portrait asks for variations of the task to reveal the correlation between positions during training.
These variations are usually achieved by multiple demonstrations.
However, demonstrations are not only time-consuming, but spatial correlations can only be correctly computed if all demonstrations are free from time misalignment, which usually demands an extra step to eliminate time warping  (e.g. \cite{abbeelAutonomousHelicopterAerobatics2010, vakanskiTrajectoryLearningRobot2012}). 
To avoid these problems, in the ball pushing task we used an artificial data augmentation procedure described as follows.

Assume a pair of trajectories consisting of robot joint angles and the respective target movements $(\bm q^{\text{demo}}, \bm x^{\text{demo}})_{1:T}$ are available  from a single demonstration of $T$ time steps.
Using the known kinematics of the robot (no dynamical models are required), we simulate $N$ spatial variations by perturbing the target with $\bm x'_{t,n} = \bm x^{\text{demo}}_{t} + \epsilon$ where $\epsilon$ is zero-mean Gaussian noise.
The respective robot pose is found by using inverse kinematics on the perturbed targets where 
$\bm q'_{t,n} \leftarrow \text{IK}(\bm x'_{t,n}, \bm q^{\text{demo}}_t)$.
The mean pose $\bm q^{\text{demo}}_t$ is used as the initial guess for the IK solver in an attempt to obtain similar joint configurations and thus preserve the normal distribution also in joint space---note that is is a heuristic for which no guarantees can be made, but in practice has worked well.
This process is repeated for each time step along the recorded nominal trajectory, providing a training set of $N$ sample variations of length $T$, such that $\{(\bm q',\bm x')_{1:N}\}_{1:T}$.
The procedure described in Section \ref{sec:PPP} can then be applied to estimate the distribution parameters of the PPMP.

\subsection*{Experimental Setup}
We used the upper body of a human-sized hydraulic humanoid comprised of seven DoFs in each arm, and three DoFs on the waist (details in \cite{cheng2007cb}).
The robot is position-controlled at the joint level, with PD controllers running at $500$ Hz on a desktop computer (Intel Core i7-4790) using Ubuntu 16.04 with a patched real-time kernel.
For simulations, we implemented a kinematic model of the robot together with a dynamic model of a ball swinging like a pendulum.

The estimation of the position of the human hand in the handover experiment and of the ball position in the pushing experiment was done by color segmentation using OpenCV methods.
The perception routines were run on the same control loop of the PPMP implemented in Python 2.7.
The PPMP runs on a conventional Ubuntu 18.04 laptop (Intel Core i7-7700HQ).
The joint angles output by the PPMP were transmitted asynchronously to the real-time controller using socket communication.

The effective frequency of joint updates was limited by the RBG-D camera in use.
When using a  Kinect2 camera, the planning frequency was approximately 30 Hz.
When using a Real Sense D450 camera, the planning frequency was approximately between 45 to 60 Hz.
For the quantitative evaluation of the performance, only the Kinect2 camera was used. 
The Kinect2 was positioned externally, at the side of the robot.
In the accompanying video, it is also possible to see a version of the experiment under the robot's point of view when the D450 was used and mounted on the robot's head.

\end{document}